\documentclass{article}

\PassOptionsToPackage{numbers, compress}{natbib}
\usepackage[main, final]{neurips_2026}
\usepackage[utf8]{inputenc} % allow utf-8 input
\usepackage[T1]{fontenc}    % use 8-bit T1 fonts
\usepackage{hyperref}       % hyperlinks
\usepackage{url}            % simple URL typesetting
\usepackage{booktabs}       % professional-quality tables
\usepackage{amsfonts}       % blackboard math symbols
\usepackage{nicefrac}       % compact symbols for 1/2, etc.
\usepackage{microtype}      % microtypography
\usepackage{xcolor}         % colors
\usepackage{graphicx}
\usepackage{amsmath}
\usepackage{amsthm}
\usepackage{pifont}
\usepackage{multirow}
\usepackage{makecell}
\usepackage{caption}
\usepackage{algorithmic}
\usepackage{algorithm}
\usepackage{diagbox}
\usepackage{subcaption}
\title{CoLaX: Context-Aware Local Explanations for Time Series Classification}

\author{%
  \textbf{Kexuan Zhang\(^1\), Paolo Frazzetto\(^2\), Xiaobei Zou\(^1\),}\\
  \textbf{Gustau Camps-Valls\(^{2\dagger}\), Yang Tang\(^{1\dagger}\)}\\[0.5em]
  {\normalsize \(^1\)East China University of Science and Technology, China}\\
  {\normalsize \(^2\)Image Processing Laboratory, Universitat de Val\`encia, Spain}\\[0.3em]
  {\small \(\dagger\) Corresponding authors}
}

\begin{document}

\maketitle

\begin{abstract}
Post-hoc explanations for time series classifiers usually score local patterns, such as subsequences or shapelets, without regard to the rest of the sequence. Yet the same motif can carry different predictive meaning under different global temporal conditions, such as trends or regimes. We propose CoLaX, a context-aware post-hoc explanation framework that builds on shapelet-driven Shapley attribution and conditions local-pattern relevance on a learned representation of the full sequence. Global context acts at two
levels: it produces sample-specific prototype weights and a temporally resolved signal that reweights the local activation map. The resulting context-conditioned segments are the players of a Shapley attribution game whose values come only from the frozen black-box classifier. On synthetic benchmarks where identical motifs change relevance across contexts, CoLaX improves AUPRC over the strongest baseline by 30\% on average. On real-world data, it achieves the highest AUPRC on ECG. Additional ablations indicate that the gains come from input-dependent conditioning rather than from additional learnable weights. CoLaX is therefore most useful when recurring local motifs change relevance across sequence-level contexts.
\end{abstract}

\section{Introduction}
Time series are sequences of indexed data that follow a specific order, and their analysis is relevant across a wide range of domains, such as healthcare \cite{puri2022forecasting} or the environment \cite{gomez2017cloud}. Although Deep Learning methods specifically designed for sequential data have a long tradition \cite{hochreiter1997long}, recent advances have demonstrated the capabilities of these technologies as highly accurate predictive tools across an increasing number of time series classification and regression tasks \cite{mohammadi2024deep}. Such models are optimized to learn high-dimensional representations from rich, raw data; yet their complexity prevents an understanding of the logic behind their predictions. This hinders the adoption of Artificial Intelligence (AI) in real-world scenarios where users need to trust its predictions \cite{dramsch2025explainability}.

This lack of transparency necessitates a paradigm shift toward time series classifiers that are inherently interpretable or post-hoc explainable and, therefore, numerous solutions have been proposed \cite{theissler2022explainable,pellicer2025explainable}.
Most existing approaches attribute predictions to individual timesteps or short segments using perturbation-based techniques \cite{turbe2023EvaluationPosthocInterpretability} or gradient-based saliency \cite{NEURIPS2020_47a3893c}. A growing line of work preserves temporal structure in the explanation itself: LIMESegment builds segment-level perturbation neighbourhoods \cite{sivill2022limesegment}, TIMING incorporates temporal dependence into integrated gradients \cite{jang2025timing}, InteDisUX combines consecutive attribution scores into discriminative segments \cite{tran2025intedisux}, and subsequence-level explainers such as Implet and ShapeX attribute predictions to meaningful motifs \cite{meng2025implet,huang2025shapex}. Together with Shapley- and shapelet-based models \cite{qu2024cnn,le2024shapeformer}, these works reflect a consensus that short and discriminative patterns
are central to both performance and interpretability.

\begin{figure}
    \centering
    \includegraphics[width=0.8\linewidth]{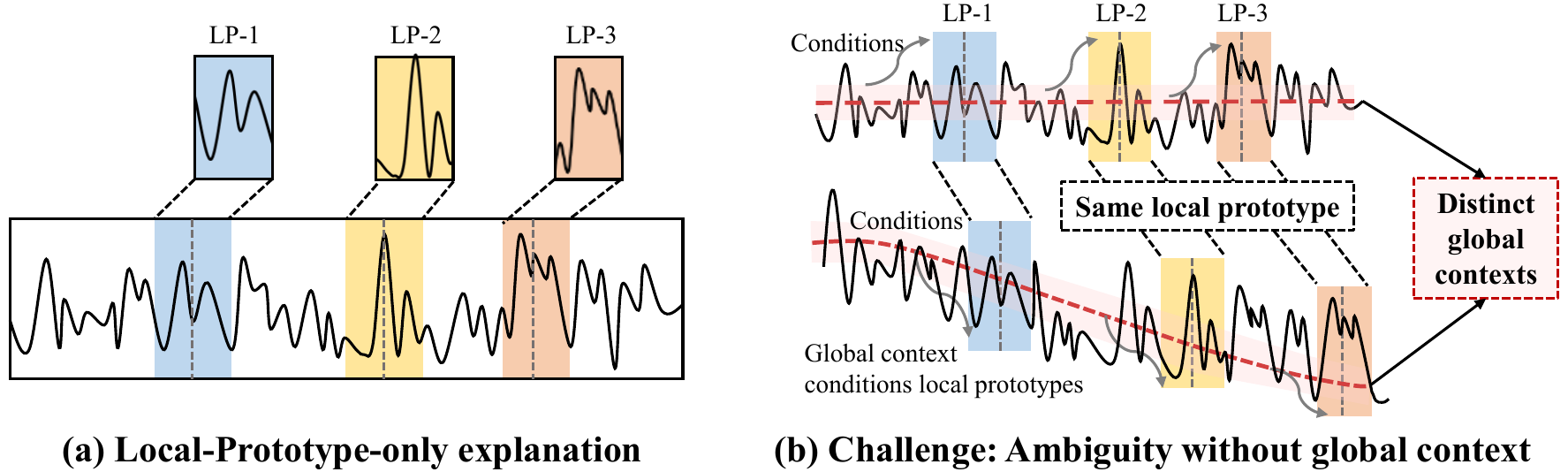}
    \caption{Ambiguity of local prototype explanations without global temporal context. (a) \textbf{Local-pattern–only explanation.} Local prototypes (LP-1 to LP-3) capture discriminative motifs but implicitly assume that their contributions are invariant across contexts. (b) \textbf{Context-dependent ambiguity under distinct global temporal structures.} Identical local patterns may appear under distinct global temporal structures. Their contributions should be interpreted as conditional on the global context, motivating the need for context-aware explanation frameworks.}
    \label{motivation}
\end{figure}

Despite these advances, existing explainers score each local pattern largely independently of the sequence in which it occurs. A local activation alone does not reveal whether a pattern is predictive in a given sequence, as identical motifs may contribute differently depending on the surrounding temporal context, such as long-term trends or regimes. Separating local and global temporal components is well established in forecasting \cite{pmlr-v162-zhou22g,wang2024rethinking} and in explainable
representation learning \cite{ronco2023role}; however, to the best of our knowledge, classification explainers do not model the relevance of a local motif as conditional on a representation of the full sequence. As illustrated in Figure \ref{motivation}, global temporal context acts as a conditioning variable that modulates the contribution of local patterns. For instance, a short-term fluctuation may signal an anomaly under a stable trend but represent normal variability within a volatile sequence (e.g., a heartbeat spike may mean something different during exercise vs. rest). While existing methods may correctly identify salient local patterns, they often fail to convey the overall temporal conditions under which the model interprets them. This gap substantially limits interpretability when predictions depend jointly on short-term dynamics and long-range structure.

In this work, we argue that faithful and informative explanations for time series classification models must jointly account for local pattern-level evidence and global temporal context. Therefore, we propose \textbf{CoLaX} (\textbf{Co}ntext-aware \textbf{L}oc\textbf{a}l e\textbf{X}planations), a context-aware post-hoc explanation framework. that explicitly constructs distinct representations for short-term local temporal patterns and long-range temporal structures, integrating them through context-dependent modulation. The global temporal context summarizes holistic sequence-level characteristics and provides the conditioning background for interpreting local pattern contributions.

Specifically, CoLaX first identifies salient local temporal patterns that contribute to frozen black-box model predictions, and then contextualizes their contributions using a global temporal representation that captures diverse temporal regimes. In this manner, the global context modulates the influence of these local explanations, revealing not only \emph{which patterns matter}, but also \emph{under which temporal contexts they matter}. This design yields structured, instance-level explanations that aim to more closely reflect the model’s inference process.

%% PAOLO VERSION 
%CoLaX builds on the shapelet-driven paradigm of ShapeX \cite{huang2025shapex}: it learns a dictionary of local prototypes, extracts the subsequences they match, and attributes predictions to these segments with Shapley values computed on the frozen black-box classifier. We do not claim these components as novel. Our contribution is to make local-pattern relevance conditional on a learned representation of the full sequence: global context produces sample-specific prototype weights and a temporally resolved signal that modulates the local activation map before segments are extracted.

%As a result, the same motif may be retained, suppressed, or reweighted as an explanatory unit depending on the sequence-level context of each input, turning a context-invariant segment attribution into an input-conditioned attribution game. Global context determines \emph{which} local occurrences are evaluated, whereas their importance is determined solely by the predictions of the black-box classifier. The resulting explanations remain instance-level; CoLaX does not provide global or cohort-level summaries.

The main contributions can be summarized as follows:
\begin{itemize}
    \item We frame time series explanation as a problem of context-dependent local relevance, where the contribution of a recurring motif may change with the sequence-level context in which it occurs;
    \item We introduce CoLaX,, which conditions local patterns on global temporal context at both the sequence level (prototype weights) and the temporal level (context signal), so that the units of segment-level Shapley attribution depend on each input, while attribution values are computed only from the frozen black-box classifier;
    \item Extensive experiments on diverse time series benchmarks demonstrate that CoLaX produces more faithful explanations than existing explainability methods, showcasing the validity of this approach when repeated local motifs shift in relevance across sequence-level contexts.
\end{itemize}

\section{Method}
%method要围绕实际属性来写！
\begin{figure*}[!ht]
    \centering
    \includegraphics[width=0.8\linewidth]{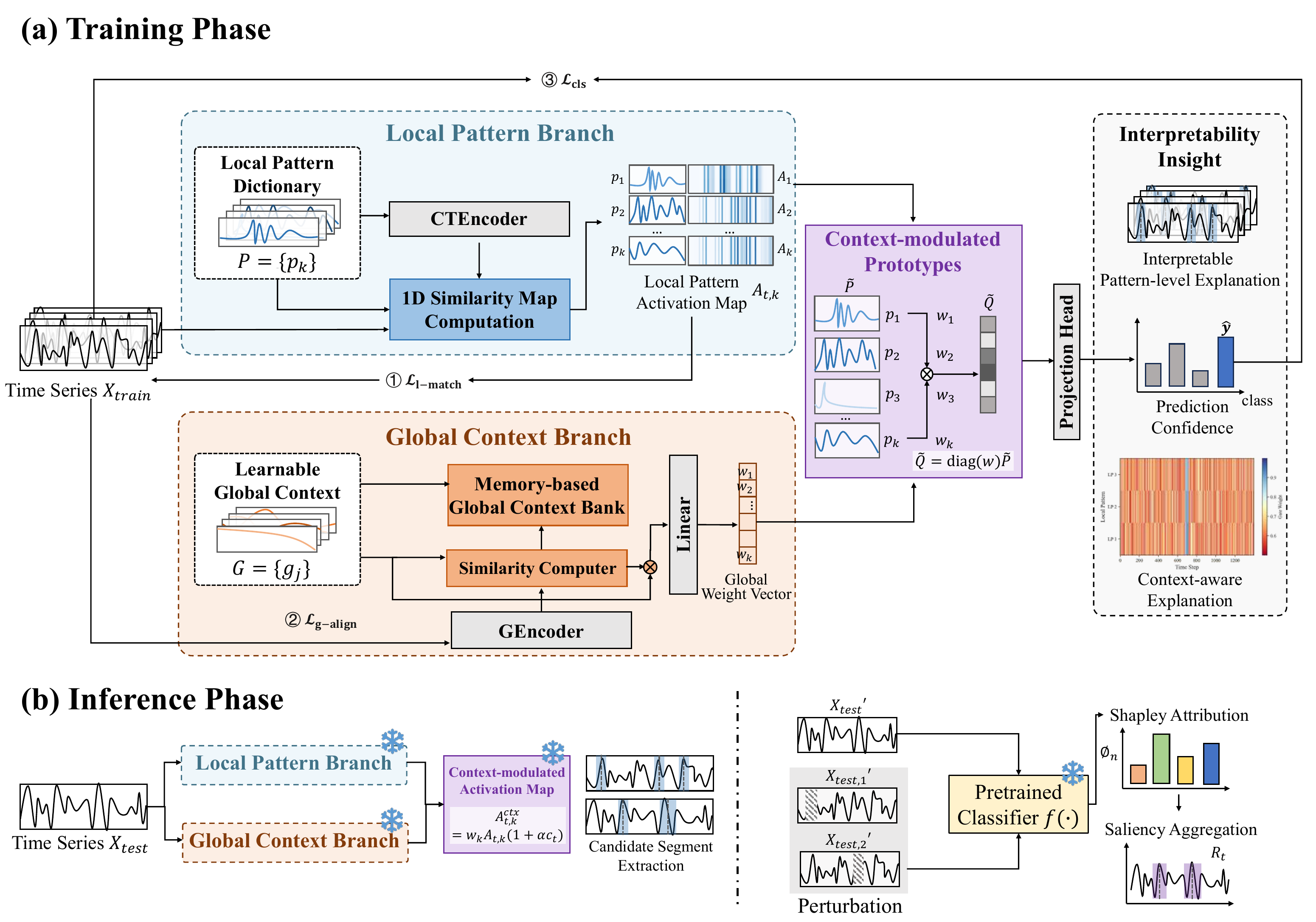}
    \caption{Overview of the proposed CoLaX framework for context-aware explanation in time series classification. The framework has two parts: a local pattern branch that builds a dictionary of short-term prototypes to detect important local subsequences, and a global context branch that uses a memory-based context bank to capture long-range background information. (a) During training, the branches are optimized end-to-end using three objectives: \ding{172} a local matching loss $L_{\text{l-match}}$ to align prototypes with salient subsequences; \ding{173} a global alignment loss $L_{\text{g-align}}$ to enforce coherent long-range structural representations; and \ding{174} a classification loss $L_{\text{cls}}$ for task-level supervision. (b) At inference time, the trained branches are frozen and serve as a post-hoc explainer. The framework explains a black-box time-series classifier by detecting context-dependent local patterns, generating pattern-based perturbations, and querying the classifier on original and perturbed inputs. It aggregates prediction changes into a saliency map, yielding instance-level, context-aware explanations.}
\label{structure}
\end{figure*}
% This work focuses on the time series classification task, denoting a multivariate time series as $\boldsymbol{X}\in \mathbb{R}^{T\times D}$ of length $T$ with $D$ features, where the associated label $y \in \{0,1\}^C$ lies in a $C$-dimensional label space, and $C$ denotes the number of classes. The objective of this task is to learn a classifier $f(\cdot)$ that maps $\boldsymbol{X}$ to a predicted label $\hat{y} = f(\boldsymbol{X})$. Given a trained classifier, post-hoc explainability aims to identify which temporal components of the input sequence contribute the most to the prediction. To formalize this, we define an explanation function $g(\cdot)$ that assigns saliency scores $R \in [0,1]^{T}$ to the input sequence, where each element $R_{t}$ quantifies the collective marginal contribution at time step $t$ to the predicted class probability (see also Appendix \ref{appendix-sal}). 
This work focuses on the time series classification task, denoting a multivariate time series as $\boldsymbol{X} \in \mathbb{R}^{T \times D}$ of length $T$ with $D$ features. The associated label $\boldsymbol{y} \in \{0,1\}^C$ lies in a $C$-dimensional label space, where $C$ denotes the number of classes. The objective is to learn a classifier $f(\cdot)$ that maps $\boldsymbol{X}$ to a predicted label $\hat{y} = f(\boldsymbol{X})$. Given a trained classifier, post-hoc explainability aims to identify which temporal components of the input sequence drive the prediction. To formalize this, we define an explanation function $\mathcal{E}(\cdot)$ that assigns a temporal saliency map $\boldsymbol{R} \in [0,1]^{T}$ to the input sequence, where each element $R_{t}$ quantifies the importance of time step $t$ to the predicted class probability (see also Appendix \ref{appendix-sal}).

Our approach is motivated by the observation that time series predictions are often driven by the interaction between short-term local dynamics and long-range temporal context \cite{pmlr-v162-zhou22g,wang2024rethinking}. Local temporal patterns capture transient, event-level fluctuations, while global temporal context characterizes slowly evolving trends and structural regimes that condition the interpretation of these patterns. Explaining model behavior, therefore, requires modeling both components and their interactions.

%To this end, we design a dual-branch architecture that explicitly models local and global temporal information as two complementary representations through a \emph{local pattern branch} and a \emph{global context branch}. The former discovers short-term and discriminative temporal patterns that provide localized evidence for the prediction, while the latter aggregates sequence-level information into a compact representation of long-range temporal structure. %These two branches are jointly optimized and integrated to form a unified representation for prediction.

To this end, we introduce CoLaX. It consists of three main components: \emph{(i)} a local pattern branch that discovers discriminative subsequences, \emph{(ii)} a global context branch that captures sequence-level structure, and \emph{(iii)} a gating mechanism that conditions local contributions on global context. We train these jointly, then use the learned representations to guide post-hoc perturbations of any black-box classifier~$f(\cdot)$. By using the discovered patterns as the basis for perturbation, CoLaX ensures that the interventions used to probe the black-box are temporally coherent and contextually grounded, leveraging the extracted dictionary of local temporal motifs and the memory bank of global contexts. In this fashion, the framework produces structured explanations that reveal both \emph{which patterns are important} and \emph{under which temporal conditions they are influential}. The overall architecture is illustrated in Figure \ref{structure}, while the algorithmic description is provided in the Appendix~\ref{appendix-alg}.

\subsection{Local Pattern Branch} \label{local_pattern}

Short-term fluctuations in time series often encode salient high-frequency dynamics, such as transient events or specific local variations, that are crucial for classification. These fine-scale patterns provide localized evidence that, while distinct from global trends, contribute significantly to model predictions. To capture these behaviors, the Local Pattern Branch learns a dictionary of temporal prototypes that represent characteristic local dynamics.

Formally, let $P=\{p_1, p_2, \cdots, p_{K_{\text{local}}}\}$
denote a collection of $K_{\text{local}}$ learnable local prototypes, where each $p_k \in \mathbb{R}^{L_p\times D}$ encodes a representative short-term pattern of length $L_p$. To enhance expressiveness and capture internal temporal dependencies, these prototypes are processed by a CNN-Transformer encoder (denoted as \text{CTEncoder}), yielding a set of refined feature representations $\widetilde{P} = \text{CTEncoder}(P) \in\mathbb{R}^{K_{\text{local}}\times D}$. This encoder integrates convolutional layers for local feature extraction with a Transformer module for modeling sequential dependencies \cite{nie2023a}.

Given the input $\boldsymbol{X}$, we compute its similarity with local prototypes via the convolution operator $I = \text{Conv}(\boldsymbol{X},\widetilde{P})$, in which the prototype vectors act as the convolution filters. The activation map $A\in[0,1]^{T\times K_{\text{local}}}$ is then obtained by applying the softmax function across the prototype dimension, i.e., $A_{t,k} = \operatorname{Softmax}(I_{t,:})_k$, where $A_{t,k}$ measures the alignment strength between the $k$-th prototype at time $t$.

Conceptually, the local pattern branch learns a dictionary of short-term motifs that summarize salient local dynamics. The activation $A_{t,k}$ reflects the relative matching strength of prototype $k$ at time step $t$, while the local matching objective encourages the learned prototypes to remain aligned with observed subsequences.

\subsection{Global Context Branch}

While local prototypes capture short-term fluctuations, their semantic interpretation often relies on the broader and slowly varying temporal background. Thus, the Global Context Branch is introduced to learn a  representation of this long-range structure, capturing information that is inaccessible to short observation windows and modulating local evidence accordingly.

\subsubsection{Global Context Representation}
To capture slowly evolving temporal dynamics, we employ a memory-based global context module. Let $\mathcal{G} = \{\boldsymbol{g}_1, \boldsymbol{g}_2, \dots, \boldsymbol{g}_{K_{\text{global}}}\}$ denote a set of $K_{\text{global}}$ learnable context vectors in the shared feature space $\mathbb{R}^D$, each encoding a prototypical long-range structure.

First, global embeddings $\boldsymbol{Z} = [\boldsymbol{z}_1, \boldsymbol{z}_2, \dots, \boldsymbol{z}_{T'}] \in \mathbb{R}^{T' \times D}$ are extracted from the input via a Transformer-based global encoder, $\boldsymbol{Z} = \text{GEncoder}(\boldsymbol{X})$, where $T'$ denotes the length of the embedding sequence. The relevance between each time-step embedding $\boldsymbol{z}_t$ and the context vectors is computed via a soft attention mechanism:
\begin{equation}
\label{eq-context-attn}
a_{t,k} = \frac{\exp(\boldsymbol{z}_t^\top \boldsymbol{g}_k / \tau_1)}{\sum_{j=1}^{K_{\text{global}}} \exp(\boldsymbol{z}_t^\top \boldsymbol{g}_j / \tau_1)},
\end{equation}
where $\tau_1$ is a temperature parameter. To derive a stable summary of the global structure, we aggregate the temporal embeddings into a single global representation: $\boldsymbol{E}_{\text{global}} = \text{Pool}(\boldsymbol{Z})$.

The central role of the global context is to make local-pattern relevance sample dependent. Given $\boldsymbol{E}_{\text{global}}$, we project the global representation into a query vector and the refined local prototypes into key representations:
\begin{equation}
\boldsymbol{q} = \boldsymbol{W}_q \boldsymbol{E}_{\text{global}}, \qquad \boldsymbol{K} = \widetilde{\boldsymbol{P}} \boldsymbol{W}_k,
\end{equation}
where $\boldsymbol{W}_q \in \mathbb{R}^{D \times D}$ and $\boldsymbol{W}_k \in \mathbb{R}^{D \times D}$ are learnable projections.

Let $\boldsymbol{k}_k=\boldsymbol{K}_{k,:}^{\top}\in\mathbb{R}^{D}$ denote the column-vector form of the $k$-th row of $\boldsymbol{K}$. The sample-specific relevance of prototype $k$ is:
\begin{equation}
\label{eq-gating}
w_k = \text{Sigmoid}\left( \frac{\boldsymbol{q}^\top \boldsymbol{k}_k}{\tau_2}\right),
\end{equation}
%Equivalently, the vectorized form is 
%\begin{equation} \label{eq-gating-vector} 
%\boldsymbol{w} = \operatorname{Sigmoid} \left( \frac{ \boldsymbol{K}\boldsymbol{q} }{ \tau_2 } \right) \in[0,1]^{K_{\text{local}}}. 
%\end{equation}
The gate is applied only after its values have been computed: 
\begin{equation} \label{eq-prototype-modulation} 
\widetilde{\boldsymbol{Q}} = \operatorname{diag}(\boldsymbol{w}) \widetilde{\boldsymbol{P}}. 
\end{equation} Hence, $\operatorname{diag}(\cdot)$ is used for prototype reweighting rather than for gate computation.
The gating weights $\boldsymbol{w}$ depend on the input-specific global representation, allowing the same local prototype to receive different weights across sequences. During training, these weights modulate the prototype representations used by the auxiliary prediction head. During post-hoc inference, they further condition local activation maps to construct context-dependent explanatory segments.

\subsubsection{Memory-based Context Bank}
To organize the global representation space around multiple prototypical temporal regimes, we maintain a memory bank preserving multiple prototypical regimes \cite{song2023memto}. For each context prototype $j$, its assigned temporal embeddings are summarized through the normalized aggregation:
\begin{equation}
\label{eq-normalized-z}
\bar{\boldsymbol{z}}_k = \frac{\sum_{t=1}^{T'} a_{t,k} \boldsymbol{z}_t}{\sum_{t=1}^{T'} a_{t,k} + \epsilon},
\end{equation}
where $\epsilon$ is a small constant for numerical stability. The context vectors are then updated through:
\begin{align}
\label{eq-memory-update}
\boldsymbol{\psi}_j &= \sigma \left(\boldsymbol{U}_\psi \boldsymbol{g}_j + \boldsymbol{W}_\psi \bar{\boldsymbol{z}}_j \right), \\
\boldsymbol{g}_j &\leftarrow (1 - \boldsymbol{\psi}_j) \odot \boldsymbol{g}_j + \boldsymbol{\psi}_j \odot \bar{\boldsymbol{z}}_j,
\end{align}
where $\boldsymbol{U}_\psi, \boldsymbol{W}_\psi$ are learnable projections and $\sigma(\cdot)$ denotes the sigmoid function. The normalized aggregation prevents the scale of the memory update from growing with sequence length and allows the context bank to represent multiple recurring global regimes.

\subsection{Prediction Head}
The context-modulated representation $\widetilde{\boldsymbol{Q}}$ is fed into a multilayer perceptron (MLP) prediction head. Crucially, the prediction head explicitly incorporates a sigmoid activation to map the fused representations to the label space:
\begin{equation}
\label{eq-prediction}
\hat{\boldsymbol{y}} = \sigma \left( \text{MLP}(\widetilde{\boldsymbol{Q}}) \right),
\end{equation}
where $\hat{\boldsymbol{y}} \in [0,1]^C$ denotes the predicted probabilities. 

\subsection{Training Losses}

To jointly optimize local pattern discovery, global context modeling, and task-level prediction, we employ a composite objective function comprising three complementary terms: (1) local semantic matching $\mathcal{L}_{\text{l-match}}$, (2) global semantic alignment $\mathcal{L}_{\text{g-align}}$, and (3) classification loss $\mathcal{L}_{\text{cls}}$. %Collectively, these components encourage interpretable representations while maintaining robust predictive performance.
\paragraph{(1) Local Semantic Matching.}
The local pattern branch is designed to capture representative short-term temporal motifs. We introduce a semantic matching loss, aligning each prototype with the most responsive subsequences in the input time series \cite{huang2025shapex}. The local matching objective anchors the learned prototypes to representative subsequences identified by their activation responses, encouraging correspondence
between prototype representations and observed temporal patterns.

Let $A_{t,k}$ denote the activation of prototype $p_k$ at time $t$. The maximally activated position for prototype $p_k$ is defined as
\begin{equation}
    t_k^{*} = \arg\max_{t} A_{t,k},
\end{equation}
from which a local subsequence of length $L_p$ is extracted:
\begin{equation}
\boldsymbol{s}_k = \boldsymbol{X}[t_k^{*} - \lfloor L_p/2 \rfloor : t_k^{*} + \lceil L_p/2 \rceil, : ].
\end{equation}
Let $\mathcal{S}_k$ denote the set of subsequences assigned to prototype $p_k$ across the batch. The local semantic matching loss is defined as
\begin{equation}
\mathcal{L}_{\text{l-match}}
=
\sum_{k=1}^{K_{\text{local}}}
\sum_{x \in \mathcal{S}_k}
\|x - p_k\|_F^2,
\end{equation}
which encourages each prototype to align with the representative local dynamics in its activation neighborhood.% This objective promotes specialization among prototypes and stabilizes the learning of fine-grained temporal motifs.
\paragraph{(2) Global Semantic Alignment.}
To encourage each global embedding to associate with the most relevant context representation while remaining discriminative from others, we introduce a global semantic alignment objective. For sample $i$, let $ \boldsymbol{e}^{(i)} = \boldsymbol{E}_{\text{global}}^{(i)} $ denote its pooled global representation. The most strongly associated context prototype is $
k_i^{+} = \arg\max_k \left( \boldsymbol{e}^{(i)\top} \boldsymbol{g}_k \right). $
The global alignment objective is
\begin{equation}
\mathcal{L}_{\text{g-align}} = -\frac{1}{N} \sum_{i=1}^{N} \log \frac{ \exp \left( \boldsymbol{e}^{(i)\top} \boldsymbol{g}_{k_i^{+}} / \tau_g \right) }{ \sum_{k=1}^{K_{\text{global}}} \exp \left( \boldsymbol{e}^{(i)\top} \boldsymbol{g}_{k} / \tau_g \right) },
\end{equation}
where the context vectors are $\ell_2$-normalized and $\tau_g$ is the temperature used for global alignment.
%This objective encourages coherent and discriminative organization of the global context space.
\paragraph{(3) Classification Objective.}
We adopt the standard cross-entropy loss $\mathcal{L}_{\text{cls}}$ for $C$ classes, calculated using the predicted probabilities $\hat{\boldsymbol{y}}$.

\paragraph{Overall Objective.}
The final training loss is given by
\begin{equation}
\mathcal{L}
=
\lambda_1 \mathcal{L}_{\text{l-match}}
+ \lambda_2 \mathcal{L}_{\text{g-align}}
+ \mathcal{L}_{\text{cls}},
\end{equation}
where $\lambda_1$, and $\lambda_2$ balance the influence of the prototype-specific and context-specific objectives. Collectively, these terms drive the model to learn structured, interpretable local patterns and global temporal contexts without compromising task-level predictive performance.

\subsection{Post-hoc Explainability}
\label{sec-posthoc}
After training, CoLaX is frozen and used to construct context-aware explanatory segments for a separate frozen black-box classifier $f(\cdot)$. Following the shapelet-driven perturbation paradigm \cite{huang2025shapex}, CoLaX conditions segment construction on global temporal context, while their contributions are determined by black-box prediction differences.

\paragraph{Context-conditioned Segment Construction.}
CoLaX incorporates global context at two levels. The pooled representation $\boldsymbol{E}_{\text{global}}$ generates sample-specific prototype weights $\boldsymbol{w}$, while the temporal context associations $a_{p,j}$ in Eq. \ref{eq-context-attn} provide a time-dependent conditioning signal:
\begin{equation}
\label{eq-temporal-context-signal}
c_t =
\text{Norm}\left[
\text{Interp}\left(
\{\max_k a_{s,k}\}_{s=1}^{T'}
\right)\right]_t,
\end{equation}
where $\text{Interp}$ restores the original temporal resolution and $\text{Norm}$ scales the context-association confidence to $[0,1]$.

The two levels of conditioning jointly modulate local activations:
\begin{equation}
\label{eq-context-conditioned-activation}
A^{\text{ctx}}_{t,k}
= w_k A_{t,k}(1+\alpha c_t),
\end{equation}
where $\alpha\geq0$ controls the temporal modulation strength. Here, $w_k$ adjusts prototype relevance across samples, while $c_t$ modulates activation strengths across temporal locations. Contiguous regions satisfying $A^{\text{ctx}}_{t,k}>\Omega$ are extracted as explanatory segments, forming $\mathcal{P}(\boldsymbol{X})=\{\mathcal{I}_1,\ldots,\mathcal{I}_N\}$, where $\Omega$ denotes the activation threshold.

\paragraph{Black-box Shapley Attribution.}
For each coalition $S\subseteq\mathcal{P}(\boldsymbol{X})$, we construct the perturbed input
\begin{equation}
\label{eq-perturbation}
\boldsymbol{X}_{S}
=\boldsymbol{m}_{S}\odot\boldsymbol{X}
+(1-\boldsymbol{m}_{S})\odot\boldsymbol{X}_{\text{base}},
\end{equation}
where $\boldsymbol{m}_{S}\in\{0,1\}^{T}$ retains segments in $S$ and masks candidate segments outside $S$, leaving non-candidate regions unchanged. The baseline $\boldsymbol{X}_{\text{base}}$ is generated by linear interpolation to reduce discontinuities.

Let $c^\star=\arg\max_c f_c(\boldsymbol{X})$ and $v(S)=f_{c^\star}(\boldsymbol{X}_S)$. The attribution of segment $n$ is computed as
\begin{equation}
\label{eq-shapley}
\phi_n =
\sum_{S\subseteq\mathcal{P}(\boldsymbol{X})\setminus\{n\}}
\frac{|S|!(N-|S|-1)!}{N!}
\left[v(S\cup\{n\})-v(S)\right].
\end{equation}
Thus, CoLaX constructs context-dependent attribution players, while their Shapley values are determined solely by the frozen black-box classifier.

\paragraph{Saliency Aggregation.}
The segment-level contributions are projected onto the temporal domain using maximum absolute aggregation:
\begin{equation}
\label{eq-final-aggregation}
R_t =
\frac{R_t^{\text{raw}}}
{\max_{t'}R_{t'}^{\text{raw}}+\epsilon},
\quad
R_t^{\text{raw}}=
\max_{n:t\in\mathcal{I}_n}|\phi_n|,
\end{equation}
where $R_t^{\text{raw}}=0$ if no segment covers $t$. Consequently, CoLaX constructs an input-conditioned attribution game whose explanatory units depend on global temporal context, while their importance is evaluated through black-box prediction changes.

\section{Experiments}

\subsection{Evaluation on Synthetic and Real-world Data}
To assess the effectiveness of CoLaX, we evaluate explanation quality across both synthetic and real-world datasets. We employ saliency score evaluations on datasets with ground-truth saliency scores, while conducting occlusion experiments on datasets lacking them. All experiments use pretrained Transformer-based classifiers as frozen black-box models, with access restricted to input-output \footnote{The code is available at \url{https://github.com/zkkkkkkkkkkx/CoLaX}}.
\paragraph{Datasets.} We evaluate our model using synthetic data comprising four motif-based binary classification datasets: MCC-E, MTC-H, MCC-H, and MTC-E \cite{germain2024persistence,huang2025shapex}. Notably, each sample in these datasets is annotated with ground-truth saliency to facilitate the evaluation of explanation quality. Regarding real-world data, we utilize the ECG dataset \cite{moody2001impact} with ground-truth labels and the \textcolor{black}{UCR Archive} \cite{UCRArchive2018} without ground-truth saliency annotations. In this case, we assess faithfulness directly on the frozen black-box classifiers through saliency-guided removal tests (Appendix~\ref{appendix-benchmark}).
% \vspace{-0.2cm}
\begin{table*}[!htbp]
    \centering
    \caption{Saliency evaluation on the four synthetic datasets. Best results are highlighted in bold, second-best underlined.}
    \tiny % 全局字体设置，或者在 minipage 内部单独设置
    \renewcommand{\arraystretch}{0.8} % 调整行高
        
        % 第一个子表：MCC-E 和 MCC-H
        \begin{tabular}{lcccccc}
            \toprule
            \multirow{2}{*}{} & \multirow{2}{*}{} & MCC-E & \multirow{2}{*}{} & & MCC-H\\
            \cmidrule(lr){2-4}
            \cmidrule(lr){5-7}
            Method & AUPRC & AUP & AUR & AUPRC & AUP & AUR \\
            \midrule
            IG          & $0.363\pm0.005$ & $0.483\pm0.008$ & $0.467\pm0.005$ & $0.436\pm0.006$ & $0.510\pm0.006$ & $0.579\pm0.005$ \\
            DYNAMASK    & $0.325\pm0.005$ & $0.472\pm0.010$ & $0.071\pm0.003$ & $0.351\pm0.003$ & $0.183\pm0.002$ & $0.157\pm0.003$ \\
            WINIT       & $0.168\pm0.004$ & $0.158\pm0.011$ & $0.386\pm0.007$ & $0.148\pm0.004$ & $0.097\pm0.005$ & $0.444\pm0.007$ \\
            CoRTX       & $0.476\pm0.006$ & $0.524\pm0.017$ & $0.459\pm0.016$ & $0.620\pm0.004$ & $0.536\pm0.016$ & $0.543\pm0.003$ \\
            MILLET      & $0.173\pm0.006$ & $0.218\pm0.007$ & $0.164\pm0.007$ & $0.302\pm0.005$ & $0.274\pm0.008$ & $0.295\pm0.003$ \\
            TimeX       & $0.369\pm0.004$ & $0.321\pm0.004$ & $\underline{0.631\pm0.003}$ & $0.444\pm0.003$ & $0.667\pm0.006$ & $0.365\pm0.003$ \\
            TimeX++     & $0.368\pm0.004$ & $0.500\pm0.006$ & $0.280\pm0.004$ & $0.639\pm0.004$ & $0.641\pm0.004$ & $0.488\pm0.002$ \\
            ShapeX      & $\underline{0.641\pm0.004}$ & $\underline{0.561\pm0.008}$ & $0.368\pm0.005$ & $\underline{0.811\pm0.001}$ & $\underline{0.684\pm0.005}$ & $\underline{0.743\pm0.006}$ \\
            \midrule
            CoLaX        & $\mathbf{0.933\pm0.003}$ & $\mathbf{0.764\pm 0.004}$ & $\mathbf{0.853\pm0.003}$ & $\mathbf{0.977\pm0.000}$ & $\mathbf{0.904\pm0.002}$ & $\mathbf{0.769\pm0.005}$ \\
            \bottomrule
        \end{tabular}
        
        \vspace{1mm} % 上下两个子表之间的间距
        
        % 第二个子表：MTC-E 和 MTC-H
        \begin{tabular}{lcccccc}
        
            \toprule
            \multirow{2}{*}{} & \multirow{2}{*}{} & MTC-E& \multirow{2}{*}{} & & MTC-H\\
            \cmidrule(lr){2-4}
            \cmidrule(lr){5-7}
            Method & AUPRC & AUP & AUR & AUPRC & AUP & AUR \\
            \midrule
           IG          & $0.147\pm0.001$ & $0.147\pm0.003$ & $0.540\pm0.004$ & $0.310\pm0.003$ & $0.532\pm0.006$ & $0.496\pm0.006$ \\
            DYNAMASK    & $0.139\pm0.001$ & $0.101\pm0.002$ & $0.280\pm0.004$ & $0.260\pm0.002$ & $0.496\pm0.007$ & $0.189\pm0.003$ \\
            WINIT       & $0.141\pm0.003$ & $0.091\pm0.005$ & $0.005\pm0.006$ & $0.150\pm0.004$ & $0.091\pm0.005$ & $0.462\pm0.006$ \\
            CoRTX       & $0.188\pm0.006$ & $0.175\pm0.017$ & $0.526\pm0.017$ & $0.243\pm0.010$ & $0.241\pm0.014$ & $0.543\pm0.006$ \\
            MILLET      & $0.142\pm0.006$ & $0.110\pm0.004$ & $0.155\pm0.015$ & $0.207\pm0.011$ & $0.231\pm0.003$ & $0.352\pm0.001$ \\
            TimeX       & $0.248\pm0.003$ & \underline{$0.630\pm0.007$} & ${0.094\pm0.001}$ & $0.380\pm0.002$ & \underline{$0.889\pm0.002$} & $0.154\pm0.001$ \\
            TimeX++     & $0.242\pm0.002$ & $0.491\pm0.006$ & $0.263\pm0.003$ & $0.390\pm0.003$ & $\mathbf{0.900\pm0.001}$ & $0.145\pm0.001$ \\
            ShapeX      & $\underline{0.610\pm0.005}$ & ${0.396\pm0.007}$ & \underline{$0.547\pm0.008$} & $\underline{0.679\pm0.001}$ & ${0.425\pm0.002}$ & $\underline{0.902\pm0.004}$ \\
            \midrule
            CoLaX        & $\mathbf{0.723\pm0.007}$ & $\mathbf{0.638\pm0.007}$ & $\mathbf{0.784\pm0.006}$ & $\mathbf{0.920\pm0.001}$ & $0.686\pm0.001$ & $\mathbf{0.977\pm0.000}$ \\
            \bottomrule
        \end{tabular}
        \label{table-synthetic}
\end{table*}
% \vspace{-0.2cm}
\begin{table*}[!htbp]
    \centering
    \caption{Saliency evaluation on the ECG dataset.}
    \tiny 
    \begin{tabular}{lccc}
        \toprule
        Method & AUPRC & AUP & AUR \\
        \midrule
        IG       & $0.418\pm0.001$ & $0.595\pm0.002$ & $0.320\pm0.001$ \\
        DYNAMASK & $0.328\pm0.001$ & $0.525\pm0.003$ & $0.108\pm0.008$ \\
        WINIT    & $0.305\pm0.001$ & $0.443\pm0.003$ & $0.347\pm0.001$ \\
        CoRTX    & $0.374\pm0.001$ & $0.497\pm0.002$ & $0.303\pm0.001$ \\
        MILLET   & $0.302\pm0.002$ & $0.472\pm0.006$ & $0.310\pm0.001$ \\
        TimeX    & $0.472\pm0.002$ & $0.566\pm0.003$ & $0.446\pm0.001$ \\
        TimeX++  & $0.660\pm0.001$ & $0.726\pm0.001$ & $0.460\pm0.001$ \\
        ShapeX   & $\underline{0.723\pm0.003}$ & $\mathbf{0.840\pm0.003}$ & $\mathbf{0.696\pm0.003}$ \\
        \midrule
        CoLaX     & $\mathbf{0.741\pm0.005}$ & $\underline{0.765\pm 0.005}$& $\underline{0.674\pm0.005}$\\
        \bottomrule
    \end{tabular}
    \label{table-ecg}
\end{table*}
\paragraph{Baselines.} Eight baselines, further described in Appendix \ref{appendix-benchmark}, are considered for comparison: IG \cite{sundararajan2017axiomatic}, Dynamask \cite{crabbe2021explaining}, WinIT \cite{leung2023temporal}, CoRTX \cite{chuang2023cortx}, MILLET \cite{early2024inherently}, TIMEX \cite{queen2023encoding}, TIMEX++ \cite{liu2024timex++}, and ShapeX \cite{huang2025shapex}.

\begin{figure}[!t]
    \centering
    \begin{minipage}[b]{0.49\linewidth}
        \centering
        \includegraphics[width=\linewidth]{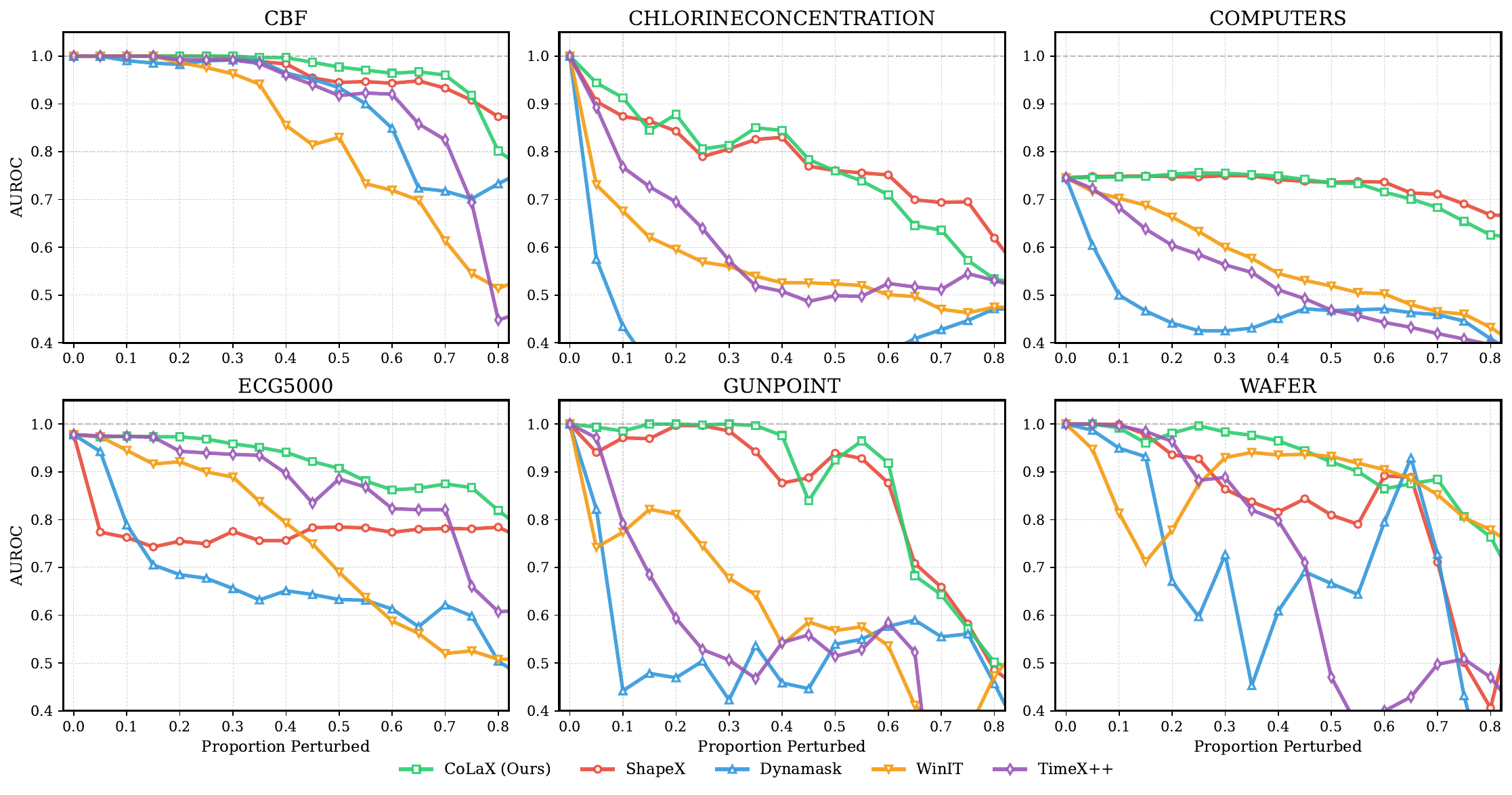}
        \caption{Least-salient-first removal on six of the 36 UCR datasets, shown as illustrative examples of different degradation behaviours.}
        \label{ucr}
    \end{minipage}
    \hfill
    \begin{minipage}[b]{0.49\linewidth}
        \centering
        \includegraphics[width=0.9\linewidth]{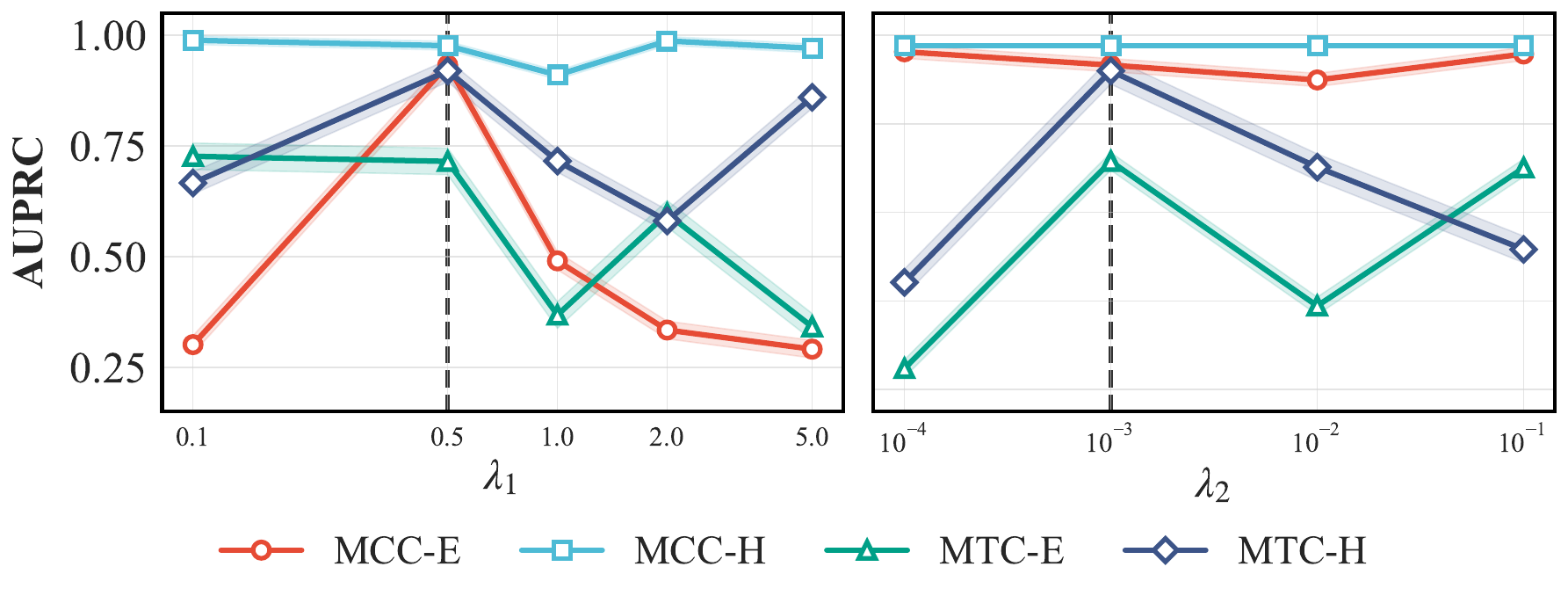}
        \caption{Sensitivity analysis on the selection of $\lambda_1$ and $\lambda_2$.}
        \label{loss_para_lambda}
    \end{minipage}
\end{figure}

\paragraph{Results.} Table \ref{table-synthetic} presents the evaluation results for the four synthetic datasets. CoLaX achieves the highest AUPRC across all four synthetic datasets and obtains the best performance on 11 of the 12 reported metrics. Regarding AUPRC, our method surpasses the second-best baseline by an average of $30.0\%$, with a peak of $45.6\%$ improvement on MCC-E. Notably, TimeX \cite{queen2023encoding} and TimeX++ \cite{liu2024timex++} exhibit degraded performance on MCC-E and MTC-E, where identical local patterns occur under different temporal contexts. In contrast, CoLaX maintains stability across both settings by conditioning local pattern contributions on the global temporal context. Results for the ECG dataset are summarized in Table~\ref{table-ecg}. CoLaX achieves the highest AUPRC, outperforming ShapeX by 2.49\%, which indicates a better overall balance between precision and recall in identifying salient regions. However, ShapeX obtains higher AUP and AUR scores, showing that CoLaX does not uniformly dominate across all saliency metrics. This indicates a favorable precision--recall trade-off rather than uniform superiority across metrics. For the heterogeneous UCR archive, we selected a representative subset spanning distinct domains and exhibiting widely varying statistical properties, as shown in Figure \ref{ucr}. The analysis reveals that CoLaX demonstrates superior performance under the least-saliency removal (bottom-k) protocol, characterized by a more graceful degradation in AUROC scores. %A more comprehensive comparison across the broader UCR archive is provided in the Appendix \ref{ucr_more}.

Further details regarding the datasets, baselines, implementation, hyperparameters, metrics, and additional results are provided in the Appendices
\ref{appendix-A} and \ref{appendix-B}. 

\begin{table*}
\centering
% \scriptsize
\caption{Ablation study of model components and training objectives. The full model is compared with variants that remove the global temporal context branch, or individual loss terms, as well as randomized control settings. All configurations are evaluated under the same experimental protocol. The results indicate that the interaction between local temporal patterns and global temporal context is critical for model performance and explanation quality.}
\tiny
\resizebox{\columnwidth}{!}{
\begin{tabular}{lcccccc}
\toprule
\multirow{2}{*}{} & \multirow{2}{*}{} & MCC-E & \multirow{2}{*}{} & & MCC-H \\
\cmidrule(lr){2-4}
\cmidrule(lr){5-7}
Method& AUPRC& AUP & AUR &AUPRC & AUP & AUR \\
\midrule
Full       & $\mathbf{0.933\pm0.003}$ & $\mathbf{0.764\pm0.004}$ & $\mathbf{0.853\pm0.003}$ & $\mathbf{0.977\pm0.000}$ & $\mathbf{0.904\pm0.002}$ & $\mathbf{0.769\pm0.005}$ \\
\emph{w/o Global}          &$0.290\pm0.004$ & $0.249\pm0.005$ & $0.320\pm0.007$ & $0.902\pm0.001$ & $0.814\pm0.002$ & $0.740\pm0.004$  \\
\emph{Random Global Context}         &$0.586\pm0.004$ & $0.529\pm0.004$ & $0.743\pm0.005$ & $0.936\pm0.000$ & $0.903\pm0.001$ & $0.768\pm0.004$  \\
\emph{Random Local Pattern}          &$0.888\pm0.003$ & $0.806\pm0.004$ & $0.641\pm0.006$ & $0.814\pm0.001$ & $0.699\pm0.002$ & $0.689\pm0.005$  \\
\emph{w/o} $\mathcal{L}_{\text{g-align}}$      & $0.586\pm0.007$ & $0.540\pm0.007$ & $0.493\pm0.007$ & $0.981\pm0.001$ & $0.863\pm0.002$ & $0.809\pm0.004$ \\
\emph{w/o} $\mathcal{L}_{\text{l-match}}$       & $0.576\pm0.004$ & $0.515\pm0.004$ & $0.750\pm0.005$ & $0.898\pm0.002$ & $0.792\pm0.004$ & $0.725\pm0.005$ \\
\bottomrule
\end{tabular}}

% \vspace{1mm}
\resizebox{\columnwidth}{!}{
\begin{tabular}{lcccccc}
\toprule
\multirow{2}{*}{} & \multirow{2}{*}{} & MTC-E & \multirow{2}{*}{} & & MTC-H \\
\cmidrule(lr){2-4}
\cmidrule(lr){5-7}
Method& AUPRC& AUP & AUR &AUPRC & AUP & AUR \\
\midrule
Full       & $\mathbf{0.723\pm0.007}$ & $\mathbf{0.638\pm0.007}$ & $\mathbf{0.784\pm0.006}$ & $\mathbf{0.920\pm0.001}$ & $\mathbf{0.686\pm0.001}$ & $\mathbf{0.977\pm0.000}$ \\
\emph{w/o Global}         & $0.237\pm0.005$ & $0.148\pm0.004$ & $0.359\pm0.009$&$0.439\pm0.007$ & $0.304\pm0.005$ & $0.596\pm0.009$   \\
\emph{Random Global Context}           &$0.627\pm0.007$ & $0.611\pm0.007$ & $0.724\pm0.006$ & $0.761\pm0.003$ & $0.611\pm0.001$ & $0.919\pm0.002$  \\
\emph{Random Local Pattern}          &$0.560\pm0.007$ & $0.495\pm0.008$ & $0.543\pm0.008$ & $0.789\pm0.002$ & $0.605\pm0.004$ & $0.910\pm0.003$  \\
\emph{w/o} $\mathcal{L}_{\text{g-align}}$      & $0.238\pm0.005$ & $0.148\pm0.004$ & $0.359\pm0.009$ & $0.439\pm0.007$ & $0.303\pm0.005$ & $0.594\pm0.009$ \\
\emph{w/o} $\mathcal{L}_{\text{l-match}}$       & $0.285\pm0.006$ & $0.224\pm0.006$ & $0.423\pm0.009$ & $0.714\pm0.005$ & $0.537\pm0.004$ & $0.888\pm0.004$ \\
\bottomrule
\end{tabular}}
\label{ablation}
\end{table*}

% \begin{figure}[!t]
%         \centering
%         % 必须加 width=\linewidth，否则图片原尺寸过大就会换行
%         \includegraphics[width=\linewidth]{figs/real_world.pdf}
%         \captionof{figure}{The performance of CoLaX on dataset from UCR.}
%         \label{ucr}
% \end{figure}
    
% \begin{figure}[!t]
%     \centering
%     \includegraphics[width=0.9\linewidth]{figs/loss_para.pdf}
%     \caption{Sensitivity analysis on the selection of $\lambda_1$ and $\lambda_2$.}
%     \label{loss_para_lambda}
% \end{figure}
\begin{figure*}[!ht]
    \centering
    \includegraphics[width=0.8\linewidth]{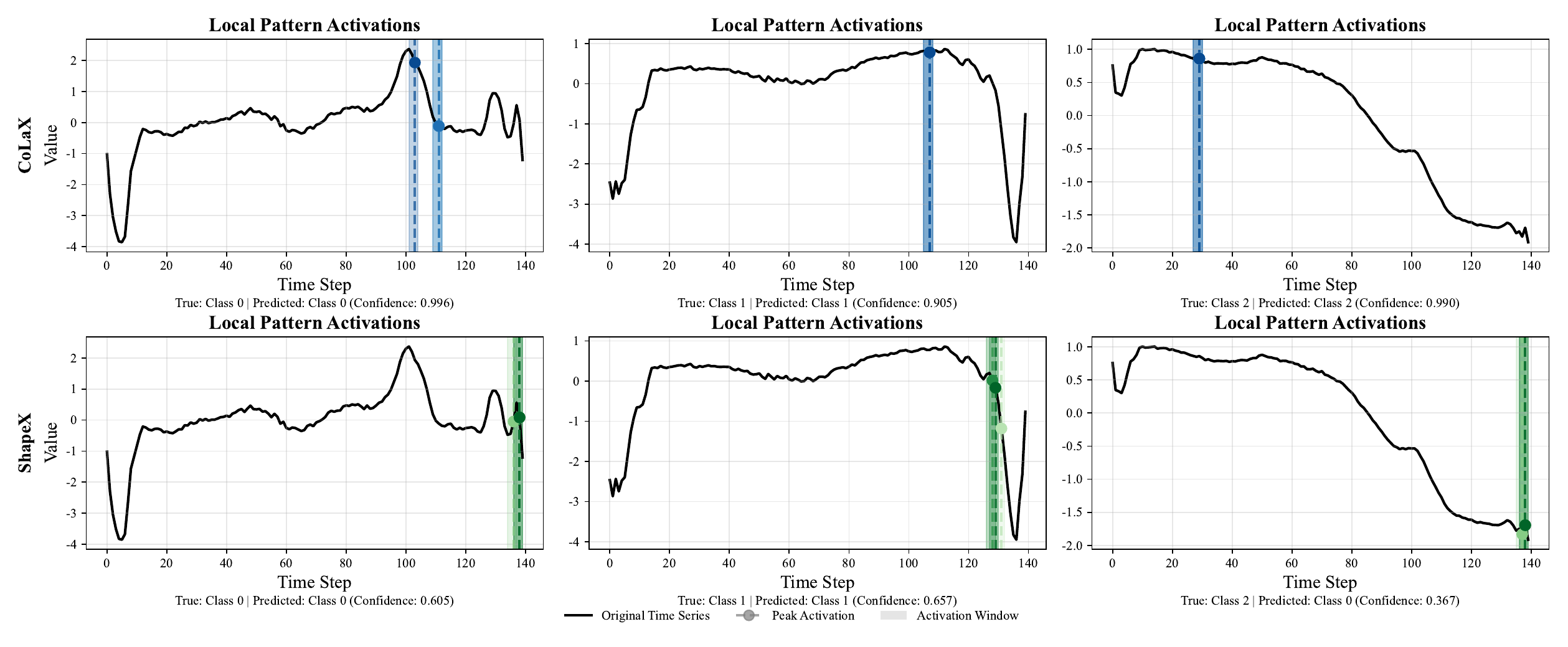}
    \caption{Representation-quality diagnostic on ECG5000: prototype activations of CoLaX (top) and ShapeX (bottom). Predictions and confidences are outputs of each method's internal surrogate head, which is used only during training; they are not predictions of the frozen black-box classifier and do not measure explanation faithfulness.}
        \label{inter}
\end{figure*}
\subsection{Ablation Study}
To assess the contribution of individual components within CoLaX, we conduct a comprehensive ablation study by systematically removing or perturbing key architectural modules and training objectives. Experiments are performed on four synthetic datasets, with the full model as the reference. Quantitative results are reported in Table \ref{ablation}. Removing the global branch (\emph{w/o Global}) causes severe degradation, with AUPRC drops of $68.9\%$ and $66.8\%$ on MCC-E and MTC-E, respectively, and an average reduction of $48.9\%$. This highlights the insufficiency of relying solely on local patterns. Replacing learned context with random signals (\emph{Random Global Context}) still yields a $17.7\%$ AUPRC drop, confirming that performance relies on semantically meaningful global context rather than mere architectural complexity. Randomizing local prototypes (\emph{Random Local Pattern}) degrades performance by $14.4\%$ on average. This asymmetry indicates that while local patterns are necessary, their interpretability depends heavily on conditioning by a coherent global context. Regarding training objectives, removing the global alignment loss (\emph{w/o} $\mathcal{L}_{\text{g-align}}$) reduces performance on several datasets, particularly MCC-E and MTC-E, although modest improvements are observed for some metrics on MCC-H, suggesting that global representations must be semantically aligned. Similarly, excluding the local matching loss (\emph{w/o} $\mathcal{L}_{\text{l-match}}$) consistently lowers performance, supporting its necessity in enforcing correspondence between prototypes and salient subsequences. To further determine whether these gains arise from sample-specific global conditioning or merely from introducing additional learnable weights over local prototypes, we replace the input-dependent prototype weights in Eq.~\ref{eq-gating} with a single learnable vector $\boldsymbol{w}_{\text{static}}\in\mathbb R^{K_\text{local}}$ shared across all samples. The global context branch and its associated alignment objective are removed, while the local
pattern branch and black-box attribution procedure are retained. CoLaX improves the average AUPRC from $0.637$ to $0.888$, outperforming the static variant on 11 of the 12 reported metrics. These results show that a fixed dataset-level weighting of prototypes is insufficient: the main benefit arises from adapting prototype relevance to the global state of each input.

% In summary, these results demonstrate that the effectiveness of CoLaX stems from the structured interaction between local patterns and global context, enabling context-aware explanations beyond unconditional local attribution.

\subsection{Parameter Analysis}
% We investigate model capacity by varying the counts of local patterns and global context (Figure \ref{para1}). Results reveal a non-additive interaction: increasing local patterns improves performance only when there is sufficient global context, confirming that global context modulates local pattern interpretation.
% % 
% Figure \ref{v-inter} demonstrates that intermediate local pattern lengths yield optimal performance. Crucially, this optimum depends on the global context, necessitating their joint optimization rather than isolated tuning.
% % 

Regarding the loss function hyperparameters, Figure \ref{loss_para_lambda} exhibits distinct peaks at our selected values ($\lambda_1=0.5, \lambda_2=10^{-3}$). Specifically, MCC-E is highly sensitive to $\lambda_1$, while the MTC datasets display an inverted U-shaped trend for $\lambda_2$, suggesting an effective balance between global alignment and regularization.

We also conducted investigations into the implementation of local patterns and global context, along with the selection of pattern length. More details can be found in Appendix \ref{appendix-para}. Our analysis infers that increasing local patterns enhances performance only in the presence of sufficient global context, confirming the synergistic relationship between global context and local patterns.

% Overall, these findings underscore that robust interpretability requires explicitly modeling the structured interaction between short-term patterns and long-range temporal context. The details will be discussed in the Appendix.

\subsection{Evaluation on Multivariate Time Series}
\label{sec:multivariate}

Although the main experiments focus on univariate time series, CoLaX is formulated for multivariate inputs $\boldsymbol{X}\in\mathbb{R}^{T\times D}$. We therefore construct a controlled two-variable benchmark by combining MCC-E and MTC-H, where four classes are defined by different cross-variable motif combinations.

We extend CoLaX to \textbf{CoLaX-M}, which learns variable-specific local prototypes through parallel branches while conditioning them on the global multivariate context. Variable-specific temporal segments are treated as candidate explanation units and jointly attributed using segment-level Shapley values on the frozen black-box classifier. For comparison, we similarly extend ShapeX to \textbf{ShapeX-M} under the same multivariate attribution protocol, together with TimeX and TimeX++.

As shown in Table \ref{tab:multivariate_synthetic}, CoLaX-M achieves the highest AUPRC and AUP, with scores of $0.535$ and $0.663$, respectively, providing initial evidence that context-dependent modulation extends effectively to multivariate attribution.

\begin{table}[!htbp]
\centering
\scriptsize
\caption{Saliency evaluation on the controlled multivariate benchmark. Best results are shown in bold and second-best results are underlined.}
\label{tab:multivariate_synthetic}
\begin{tabular}{lccc}
\toprule
Method & AUPRC $\uparrow$ & AUP $\uparrow$ & AUR $\uparrow$ \\
\midrule
TimeX
& $0.231\pm0.002$
& $0.210\pm0.004$
& $\underline{0.531\pm0.004}$ \\

TimeX++
& $0.299\pm0.001$
& $0.267\pm0.002$
& $\mathbf{0.613\pm0.006}$ \\

ShapeX-M
& $\underline{0.488\pm0.004}$
& $\underline{0.581\pm0.008}$
& $0.411\pm0.007$ \\

CoLaX-M
& $\mathbf{0.535\pm0.005}$
& $\mathbf{0.663\pm0.004}$
& $0.408\pm0.006$ \\
\bottomrule
\end{tabular}
\end{table}

\subsection{Showcase of Explanations}

To assess whether the models have learned genuine electrophysiological features, we visualize the internal surrogate representations of CoLaX and ShapeX on the ECG5000 dataset in Figure \ref{inter}. The visualization reveals a fundamental divergence in the underlying representation logic.

CoLaX (Top) demonstrates superior representation fidelity by consistently attending to clinically relevant physiological features, such as the T-wave and QRS complex, yielding high surrogate confidence. In contrast, ShapeX (Bottom) exhibits shortcut learning by fixating on tail artifacts. This reliance on boundary noise leads to a critical failure in the PVC case, where the arrhythmia is categorized as Normal because the model ignores the actual pathological deflections. Such a discrepancy underscores the critical importance of learning global morphological contexts rather than relying on discriminative local fragments. Details are provided in Appendix~\ref{appendix-ecg}.

\section{Conclusion}
%GUSTAU VERSION - SHORTER, MORE COMPELLING and AUTHORITATIVE
%This paper identified and addressed a fundamental flaw in time series interpretability: the reliance on context-agnostic local explanations. While standard methods treat temporal patterns in isolation, we demonstrated that such approaches ignore the semantic shifts caused by long-range temporal structures. To resolve this, we introduced CoLaX, a framework that treats local dynamics and global context as complementary components. By employing an explicit conditioning mechanism, CoLaX captures how global context modulates the predictive weight of local features, providing a more faithful representation of a model's decision logic. Our results establish that context-dependency is not just an auxiliary feature but a requirement for reliable and model-agnostic time-series explanation. By bridging the gap between local evidence and global context, our framework provides a robust foundation for interpretable AI, paving the way for trustworthy deployment in safety-critical scientific and engineering applications. 
This paper addressed a limitation of current time series explanations: local patterns are scored independently of the
sequence in which they occur, although the same motif may contribute differently under different global temporal conditions. We introduced CoLaX, a post-hoc explanation
framework that builds on shapelet-driven Shapley attribution and conditions local-pattern relevance on a learned representation of the full sequence, at both the sequence and the temporal level. Global context thus determines which local
occurrences are evaluated as explanatory units, whereas their importance is computed solely from the frozen black-box
classifier.

While CoLaX addresses this local-global ambiguity, it also presents certain limitations. As a post-hoc surrogate framework, it provides correlational rather than strictly causal explanations.
In the future, we will focus on deepening the causal investigation to disentangle how global context drives local pattern importance, while simultaneously extending the framework to support decisions in practical applications such as meteorological analysis and health monitoring.

\newpage
\begin{ack}
% K.Z. acknowledges support from...
This work was supported by the National Natural Science Foundation of China (Grant Nos. 62293502, 62233005, and U2441245), Shanghai Municipal Special Program for Basic Research on General AI Foundation Models (Grant No. 2025SHZDZX026D01), and the European Union's Horizon Europe research and innovation programme (Grant Agreement Nos. 101130544, ThinkingEarth, and 101120237, ELIAS). The computations in this research were performed using the CFFF platform of Fudan University. The views and opinions expressed are solely those of the authors and do not necessarily reflect those of the European Union. Neither the European Union nor the granting authority can be held responsible for them.
\end{ack}

{\small
\bibliographystyle{plainnat}
\bibliography{main_new}
}

\newpage
\appendix
% \newpage
% \makeatletter
% \setcounter{section}{0}
% \setcounter{page}{1}      
%     \begin{center}
%         \vspace*{1cm}
%         {\huge \bfseries - Appendix - \\CoLaX: Context-Aware Local Explanations for Time Series Classification \par}
%         \vspace{1.0cm}
%         {\large \bfseries Submission \#3811 \par}
%         \vspace{1.5cm}
%     \end{center}
\section{Related Work}
%\subsection{Time series explainability}
% 现有方法（参考shap）
The field of time series explainability has attracted significant attention for its role in interpreting the decision-making processes of black-box models applied to temporal data \cite{theissler2022explainable}. The inherent opacity of pure black-box models obscures their internal mechanisms, rendering decision logic difficult for users to comprehend.\\
Explainability is defined as the ability to elucidate a model's internal functioning and provide meaningful insights into how predictions are made \cite{arsenault2025survey}. Explainable Artificial Intelligence (XAI) addresses this challenge through two primary categories: intrinsic transparency and post-hoc approaches. The former utilizes simple structures or inherently interpretable formulations, achieving transparency by design. Conversely, post-hoc approaches are applied after model training to interpret what a model has learned, without altering its underlying predictive mechanism \cite{turbe2023EvaluationPosthocInterpretability}. Within this class of methods, gradient-based techniques assess input sensitivity but often fail to yield temporally coherent or stable explanations in time series settings \cite{sundararajan2017axiomatic,fortuin2018som}. Perturbation-based methods offer an alternative by estimating feature importance through the measurement of prediction changes under systematic masking strategies \cite{queen2023encoding,liu2024timex++}. Extending this paradigm, Shapley-based approaches approximate cooperative game-theoretic attributions to provide both local and global explanations for complex temporal models \cite{huang2025shapex}. As time series models grow in scale and complexity, research has increasingly shifted from point-wise attributions to structured explanations that operate at the level of subsequences or dynamical patterns \cite{yang2016shapelet,cho2021interpreting,huang2025shapex}. However, many existing approaches do not explicitly model how sequence-level temporal context conditions the selection and attribution of local explanatory patterns.\\
This perspective motivates the development of frameworks that transcend isolated saliency scores to capture the interplay between local transient patterns and global system evolution.\\
\section{Experimental Details}\label{appendix-A}
\subsection{Dataset}\label{appendix-data}
We evaluate our CoLaX on both synthetic and real-world datasets.
\paragraph{Synthetic Datasets}
As noted in \cite{huang2025shapex}, class-discriminative features in real-world applications are often defined by specific temporal patterns rather than simple amplitude differences. Consequently, we utilize the four motif-based binary classification datasets proposed in \cite{huang2025shapex}: MCC-E, MTC-H, MCC-H, and MTC-E. The statistical details are provided in Table \ref{real_world}. These datasets align more closely with real-world data-generation mechanisms, enabling us to more accurately assess whether an explanation method can effectively identify the time-series patterns crucial for classification.
\begin{table*}[th]
\centering
\caption{Overview of Synthetic Datasets}
% Resizebox is used to fit the wide table within the text width
\resizebox{\textwidth}{!}{%
\begin{tabular}{l|l|cccc|l|c}
\hline
\textbf{Category} & \textbf{Dataset Name} & \textbf{Classes} & \textbf{Testing Samples} & \textbf{Training Samples} & \textbf{Length of Series} & \multicolumn{1}{c|}{\textbf{Description}} & \textbf{True Saliency Score} \\ \hline
\multirow{4}{*}{Synthesis} & MCC-E & 2 & 2,000 & 10,000 & 800 & Shapelet count for classification, equal amplitude & Yes \\
 & MCC-H & 2 & 2,000 & 10,000 & 800 & Shapelet count for classification, higher amplitude & Yes \\
 & MTC-E & 2 & 2,000 & 10,000 & 800 & Shapelet type for classification, equal amplitude & Yes \\
 & MTC-H & 2 & 2,000 & 10,000 & 800 & Shapelet type for classification, higher amplitude & Yes \\ \hline
\end{tabular}%
}

\end{table*}
\begin{table*}[th]
\centering
\caption{Overview of Real-world Datasets}
\label{real_world}
\resizebox{\textwidth}{!}{%
\begin{tabular}{l|l|cccc|l|c}
\hline
\textbf{Category} & \textbf{Dataset Name} & \textbf{Classes} & \textbf{Testing Samples} & \textbf{Training Samples} & \textbf{Length of Series} & \multicolumn{1}{c|}{\textbf{Description}} & \textbf{True Saliency Score} \\ \hline
\multirow{37}{*}{Real-world Dataset} &ECG & 2 & 37004 & 55507 & 360 & Biomedical ECG signals with expert annotation & No \\
& ACSF1 & 10 & 100 & 100 & 1460 & Device & No \\
& ARROWHEAD & 3 & 175 & 36 & 251 & Image & No \\
& BME & 3 & 150 & 30 & 128 & Simulated & No \\
&CAR & 4 & 60 & 60 & 577 & Sensor & No \\
&CBF & 3 & 900 & 30 & 128 & Simulated & No \\
&ChlorineConcentration & 3 & 3840 & 467 & 166 & Simulated & No \\
&CinCECGTorso  & 4 & 1380 & 40 & 1639 & ECG & No \\
&Coffee  & 2 & 28 & 28 & 286 & Spectro & No \\
 & Computers & 2 & 250 & 250 & 720 & Device & No \\
 & CricketX & 12 & 390 & 390 & 300 & Har & No \\
 & CricketY & 12 & 390 & 390 & 300 & Har & No \\
 & CricketZ & 12 & 390 & 390 & 300 & Har & No \\
 & Crop & 24 & 16800 & 7200 & 46 & Image & No \\
 & DiatomSizeReduction & 4 & 306 & 16 & 345 & Image & No \\ 
 & DistalPhalanxOutlineAgeGroup & 3 & 139 & 400 & 80 & Image & No \\ 
 & DistalPhalanxOutlineCorrect & 2 & 276 & 600 & 80 & Image & No \\
 & ECG200 & 2 & 100 & 100 & 96 & ECG & No \\
 & ECG5000 & 5 & 4500 & 500 & 140 & ECG & No \\
 & FordA & 2 & 1320 & 3601 & 500 & Sensor & No \\
 & FreezerRegularTrain & 2 & 2850 & 150 & 301 & Device & No \\
 & GunPoint & 2 & 150 & 50 & 150 & Har & No \\
 & ItalyPowerDemand & 2 & 1029 & 67 & 24 & Sensor & No \\
 & Lightning7 & 7 & 73 & 70 & 319 & Sensor & No \\
 & Mallat & 8 & 2345 & 55 & 1024 & Simulated & No \\
 & MedicalImages & 10 & 760 & 381 & 99 & Image & No \\
 & ShapesAll & 60 & 600 & 600 & 512 & Image & No \\
 & SmoothSubspace & 3 & 150 & 150 & 15 & Simulated & No \\
 & SonyAIBORobotSurface1 & 2 & 601 & 20 & 70 & Sensor & No \\
 & Strawberry & 2 & 370 & 613 & 235 & Spectro & No \\
 & Symbols & 6 & 995 & 25 & 398 & Image & No \\
 & SyntheticControl & 6 & 300 & 300 & 60 & Simulated & No \\
 & TwoPatterns & 4 & 4000 & 1000 & 128 & Simulated & No \\
 & UMD & 3 & 144 & 36 & 150 & Simulated & No \\
 & UWaveGestureLibraryAll & 8 & 896 & 3582 & 945 & Har & No \\
 & WAFER & 2 & 6164 & 1000 & 152 & Sensor & No \\
 & WORMS & 5 & 77 & 181 & 900 & Motion & No \\
 \hline
\end{tabular}%
}
\end{table*}
\paragraph{Real-world dataset.}
We evaluate CoLaX on the ECG dataset following the experimental protocol of ShapeX~\cite{huang2025shapex}. This benchmark is used for binary atrial fibrillation (AF) classification and consists of 55,507 training samples and 37,004 test samples, each containing 360 time steps. Following ShapeX, we use the same ECG data and reference saliency annotations for explanation evaluation. These reference regions serve as domain-informed proxies rather than direct human explanations of the black-box predictions. We additionally evaluate CoLaX on selected datasets from the UCR Time Series Classification Archive.

\subsection{Data Assets and Licenses}
All datasets used in this study are publicly accessible research benchmarks. The four synthetic motif-based datasets (MCC-E, MCC-H, MTC-E, and MTC-H) follow the data generation protocol of ShapeX \cite{huang2025shapex}, and are used with appropriate attribution to the original authors. The MIT-BIH Arrhythmia Database is available through PhysioNet under the Open Data Commons Attribution
License (ODC-By 1.0). The UCR Time Series Classification Archive is publicly accessible, with usage and redistribution conditions varying across individual datasets. We acknowledge the original sources and follow their applicable terms of use.

\subsection{Benchmarks}\label{appendix-benchmark}
We evaluate explanation quality using a diverse set of benchmark methods developed for interpreting deep learning models on time series data, covering gradient-based, perturbation-based, surrogate-model, and inherently interpretable approaches.
\paragraph{Integrated Gradients (IG)} \cite{sundararajan2017axiomatic}.
Integrated Gradients is a classical gradient-based attribution method that explains model predictions by integrating gradients along a straight path from a baseline input to the original input. By accumulating the sensitivity of the output with respect to each input dimension, IG assigns attribution scores that satisfy desirable axiomatic properties such as completeness and sensitivity.
\paragraph{Dynamask} \cite{crabbe2021explaining}.
Dynamask is a perturbation-based explainer specifically designed for time series data. It learns a continuous mask that selectively occludes parts of the input while minimizing distortion to the model’s prediction. The mask gradually deforms the input toward a carefully chosen baseline, yielding temporally coherent saliency patterns.
\paragraph{WinIT} \cite{leung2023temporal}.
WinIT is a feature-removal-based explanation method that evaluates the importance of temporal segments by measuring the change in prediction induced by their removal. By eliminating specific time intervals and modeling dependencies across time steps, WinIT captures both local relevance and temporal interactions.
\paragraph{CoRTX} \cite{chuang2023cortx}.
CoRTX adopts an augmentation-based contrastive learning strategy to efficiently approximate Shapley values. By leveraging a small number of contrastive perturbations, it estimates feature attributions with reduced computational overhead while retaining the theoretical grounding of SHAP.
\paragraph{MILLET} \cite{early2024inherently}.
MILLET provides inherent interpretability for time series classifiers through a Multiple Instance Learning (MIL) formulation. It produces sparse and instance-level explanations directly from the model architecture, achieving interpretability without relying on post-hoc perturbations or sacrificing predictive performance.
\paragraph{TIMEX} \cite{queen2023encoding}.
TIMEX generates explanations by training an interpretable surrogate model to mimic a pretrained black-box classifier. By preserving latent space relationships between inputs and predictions, TIMEX delivers structured and faithful explanations that reflect the behavior of the original model.
\paragraph{TIMEX++} \cite{liu2024timex++}.
TIMEX++ extends TIMEX by incorporating an information bottleneck principle. It produces label-preserving, in-distribution explanations through constrained input perturbations, explicitly mitigating trivial solutions and distributional shift issues commonly encountered in perturbation-based methods.
\paragraph{ShapeX} \cite{huang2025shapex}.
ShapeX explains time series classifiers by segmenting inputs into semantically meaningful, shapelet-driven subsequences and assigning saliency scores via Shapley value attribution. This segment-level formulation enables explanations that align with the model's underlying decision logic and capture higher-level temporal structures.
\paragraph{Metrics}
Following \cite{queen2023encoding}, we adopt AUP (Area Under the Precision Curve), AUR (Area Under Recall Curve), and AUPRC (Area Under the Precision–Recall Curve) to quantitatively evaluate saliency maps in experiments where ground-truth explanations are available. These metrics assess the discrepancy between the generated saliency scores and the true salient regions. In all cases, metric values closer to 1 indicate higher agreement with the ground truth and thus higher explanation quality.

In occlusion-based experiments, where ground-truth saliency annotations are unavailable, we evaluate explanation faithfulness using the AUROC (Area Under the Receiver Operating Characteristic Curve) of the underlying black-box classifier. Specifically, AUROC measures the degradation in classification performance as progressively least salient (bottom-k) subsequences identified by the explanation method are removed. In this setting, a faithful explainer should result in a slower and smoother performance drop, as it successfully prioritizes the removal of non-informative background regions while preserving key motifs.

Let $x \in \mathbb{R}^{T\times D}$ denote an input time series, where $T$ is the temporal length and $D$ the feature dimension.
Let $Q$ be the ground-truth saliency matrix, where $Q_{t}=1$ indicates that time step $t$ is truly salient, and 0 otherwise. Let $R \in [0,1]^T$ denote the saliency map produced by the explanation method. Given a detection threshold $\tau \in (0,1)$,  the binary saliency estimator $\hat{Q}_{t}(\tau)$ is defined as:
\begin{equation}
    \hat{Q}_{t}(\tau)=\left\{
\begin{aligned}
&1\qquad \text{if}\quad R_{t}\geq  \tau\\
&0\qquad \text{else}. \\
\end{aligned}
\right.
\end{equation}
We define the set of truly salient indices and the set predicted as salient by the explanation method as:
\begin{align*}
A &= \{t \in [1 : T] \mid Q_{t} = 1\} \\
\hat{A}(\tau) &= \{t \in [1 : T] \mid \hat{Q}_{t}(\tau) = 1\} .
\end{align*}
For each threshold $\tau$, the precision and recall are defined as:
\begin{align*}
\mathbf{P}(\tau) = \frac{|A \cap \hat{A}(\tau)|}{|\hat{A}(\tau)|} \\
\quad \mathbf{R}(\tau) = \frac{|A \cap \hat{A}(\tau)|}{|A|}.
\end{align*}
The Area Under the Precision Curve (AUP) and Area Under Recall Curve (AUR) are computed as:
\begin{align*}
\text{AUP} &= \int_0^1 \mathbf{P}(\tau) d\tau \\
\text{AUR} &= \int_0^1 \mathbf{R}(\tau) d\tau .
\end{align*}
To jointly assess the trade-off between precision and recall, we additionally compute the Area Under the Precision–Recall Curve (AUPRC):
\begin{equation}
\text{AUPRC} = \int_{0}^{1} \mathbf{P}(\mathbf{R})d\mathbf{R},
\end{equation}
where precision is treated as a function of recall as the threshold $\tau$ varies.

\subsubsection{Experimental Settings} \label{setting}
All experiments were conducted on an NVIDIA A100 GPU with all models (our framework and baselines) implemented in PyTorch. For each dataset, we use the standard train/test split from the corresponding benchmark and reserve 20\% of the training set for validation to tune hyperparameters and perform early stopping. Black-box classifiers are trained with cross-entropy loss until convergence, using validation accuracy for early stopping. Unless otherwise specified, we use each baseline's default hyperparameters; the main settings for all datasets are summarized in Table~\ref{imp} and kept fixed to ensure fair comparison. Results are averaged over five random seeds (controlling initialization and data shuffling), and we report the mean and standard deviation (shown as error bars where applicable).

\begin{table}[]
    \centering
    \caption{Training parameters across different datasets.}
    % Resizebox is used to fit the wide table within the text width
    \scriptsize 
    \begin{tabular}{c|c|c|c|c|c}
        \toprule
         Parameter& MCCE&MCCH&MTCE&MTCH&ECG \\
         \midrule
         Length of pattern&100&100&100&100&40 \\
         Number of local pattern&2&2&2&2&2\\
         Number of global context&2&2&2&2&2\\
         Number of CTEncoder&6&6&6&1&1\\
         Number of GEncoder&1&6&1&1&1\\
         \bottomrule
    \end{tabular}
    
    \label{imp}
\end{table}
\section{Further Analysis Experiments}\label{appendix-B}
\subsection{Computational Cost} \label{cost}
To assess the computational cost of each baseline and our CoLaX, we compare the inference time required for explanation generation across several datasets, as shown in Table \ref{compute cost}. As observed, our method maintains a competitive computational profile. It is significantly more efficient than \textcolor{black}{DYNAMASK} and demonstrates efficiency comparable to ShapeX. While it is well known that computing Shapley values is highly time-consuming, and CoLaX may not be the fastest method among all baselines, it strikes a superior balance between efficiency and performance. It keeps the computational overhead within an acceptable range while achieving significantly higher accuracy.

\begin{table}[]
    \centering
    \caption{Comparison of wall-clock time (in seconds) on different datasets.}
    \scriptsize
    \begin{tabular}{c|c|c|c|c|c}
        \toprule
         \diagbox{Dataset}{Method}&Ours&ShapeX&\textcolor{black}{DYNAMASK}&WINIT&TimeX++\\
         \midrule
         MCC-E&68.1&58.3&385.2&43.4&20.9\\
         MTC-E&72.1&63.1&383.8&30.7&27.2\\
         MCC-H&70.3&58.6&384.3&39.7&23.0\\
         MTC-H&55.4&51.4&383.1&39.7&21.2\\
         ECG5000&35.3&33.1&181.8&8.7&17.6\\
         \bottomrule
    \end{tabular}
    
    \label{compute cost}
\end{table}

\subsection{Parameter Analysis} \label{appendix-para}
\begin{figure}
    \centering
    \includegraphics[width=0.7\linewidth]{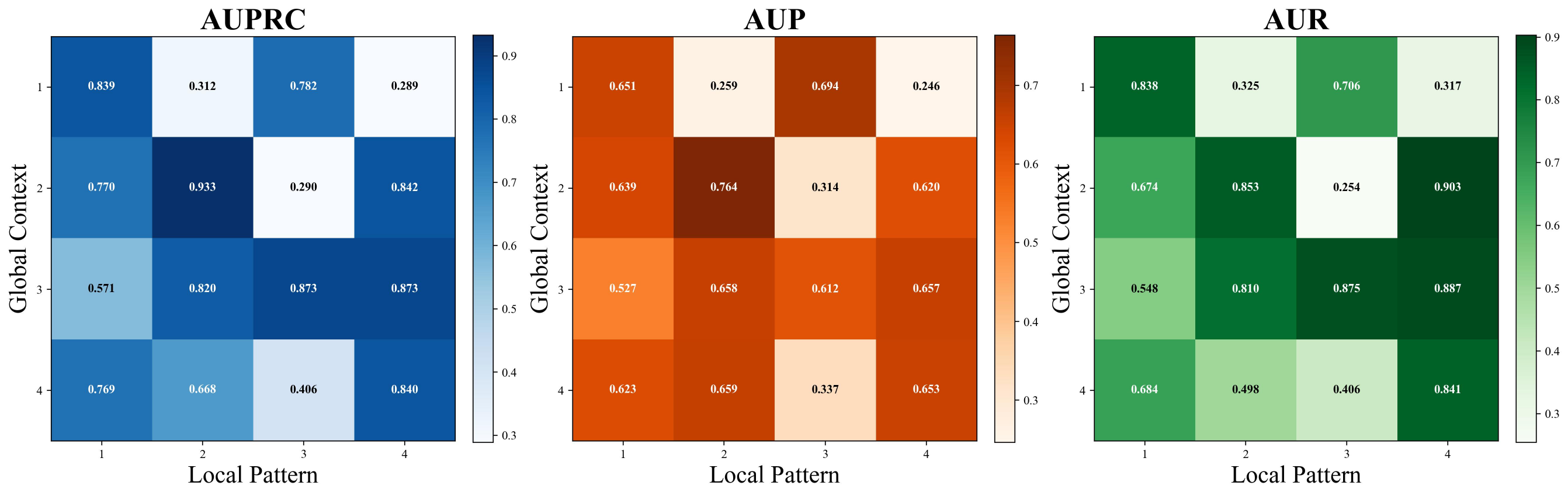}
    \caption{Metric under different local–global interaction configurations on MCC-E datasets.}
    \label{para_mcce}
\end{figure}
\begin{figure}
    \centering
    \includegraphics[width=0.7\linewidth]{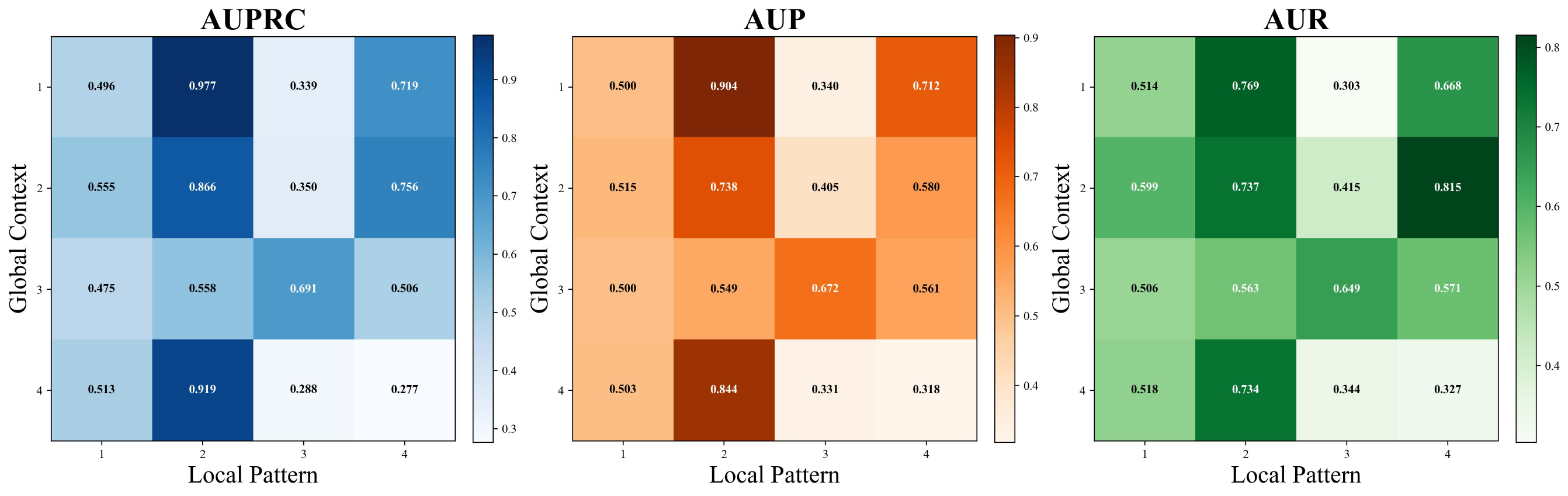}
    \caption{Metric under different local–global interaction configurations on MCC-H datasets.}
    \label{para_mcch}
\end{figure}
\begin{figure}
    \centering
    \includegraphics[width=0.7\linewidth]{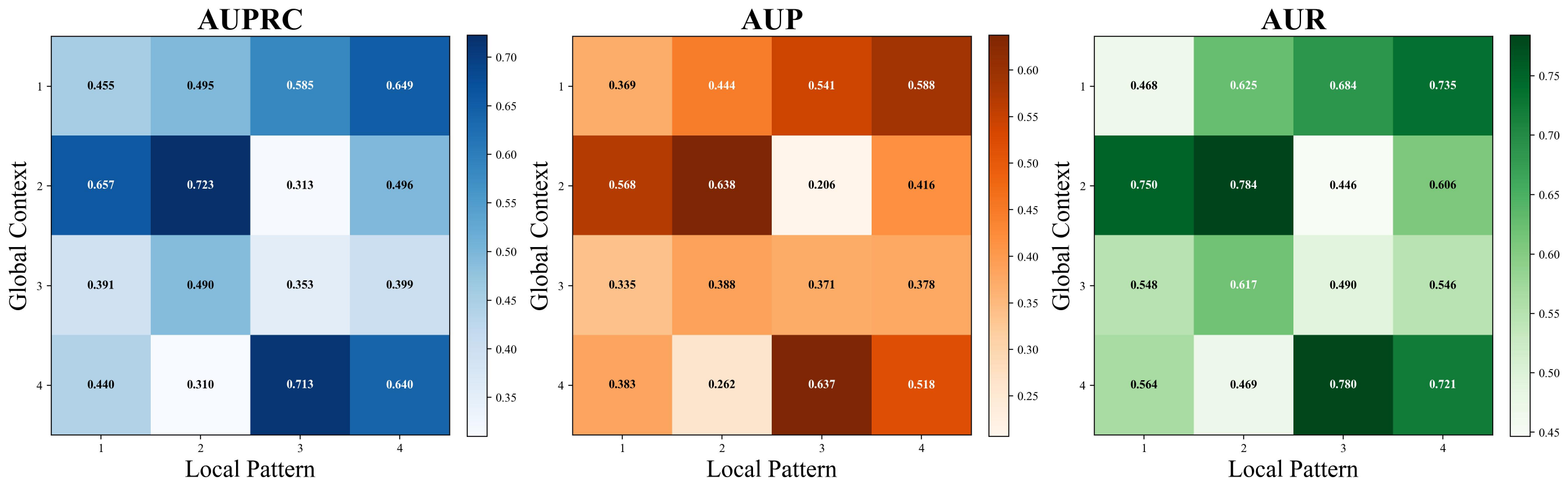}
    \caption{Metric under different local–global interaction configurations on MTC-E datasets.}
    \label{para_mtce}
\end{figure}
We investigate model capacity by varying the counts of local patterns and global context, as shown in Figure \ref{para_mcce}, \ref{para_mcch}, and \ref{para_mtce}. Results reveal a non-additive interaction: increasing local patterns improves performance only when there is sufficient global context, confirming that global context modulates local pattern interpretation.

\begin{figure*}[t] % 建议使用 [t] 让图浮动到页首，或者 [htbp]
    \centering
    % --- 左图 Minipage ---
    \begin{minipage}[c]{0.48\linewidth} % [b] 表示底部对齐，保证 Caption 对齐
        \centering
        % 注意：放入 minipage 后，width 通常设为 \linewidth (即占满当前小容器的宽度)
        \includegraphics[width=0.8\linewidth]{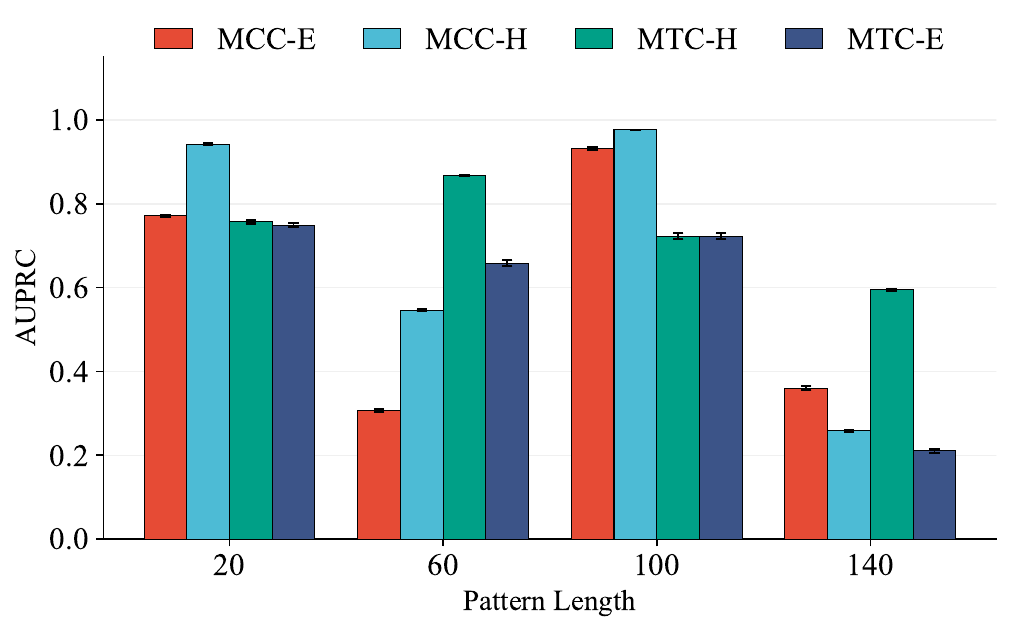}
        \caption{AUPRC under different pattern length configurations on synthetic datasets.}
        \label{v-inter}
    \end{minipage}
    \hfill % 在两个 minipage 之间添加弹性空白，使它们撑满一行
    % --- 右图 Minipage ---
    \begin{minipage}[c]{0.48\linewidth}
        \centering
        \includegraphics[width=\linewidth]{figs/loss_para.pdf}
        \caption{Analysis on the selection of $\lambda_1$ and $\lambda_2$.}
        \label{loss_para}
    \end{minipage}
\end{figure*}

Figure \ref{v-inter} demonstrates that intermediate local pattern lengths also yield optimal performance. Crucially, this optimum depends on the global context, necessitating their joint optimization rather than isolated tuning.

Regarding the loss function hyperparameters, Figure \ref{loss_para} exhibits distinct peaks at our selected values ($\lambda_1=0.5, \lambda_2=10^{-3}$). Specifically, MCC-E is highly sensitive to $\lambda_1$, while the MTC datasets display an inverted U-shaped trend for $\lambda_2$, suggesting an effective balance between global alignment and regularization.

Overall, these results indicate that the numbers of local patterns and global contexts, and the pattern length, should be
tuned jointly rather than in isolation.

\subsection{Robustness Evaluation}
\label{appendix-robustness}

We evaluate the local stability of CoLaX explanations under three types of input perturbations: Gaussian noise ($\sigma \in \{0.005, 0.01, 0.02, 0.05, 0.1\}$), temporal shifts ($s \in \{1, 2, 3, 5\}$), and amplitude scaling ($a \in \{0.8, 0.9, 1.1, 1.2\}$). Only perturbations preserving the frozen black-box classifier's predicted label are retained. Explanation consistency is measured using Spearman rank correlation ($\rho$) and top-$k$ intersection-over-union (IoU), with higher values indicating greater stability.

\begin{table*}[t]
\centering
\caption{Robustness of CoLaX explanations under prediction-preserving perturbations. Results are averaged across perturbation strengths. Higher is better.}
\label{tab:robustness}
\setlength{\tabcolsep}{6pt}
\renewcommand{\arraystretch}{1.1}
\scriptsize
\begin{tabular}{lcccccc}
\hline
\multirow{2}{*}{Dataset}
& \multicolumn{2}{c}{Gaussian Noise}
& \multicolumn{2}{c}{Temporal Shift}
& \multicolumn{2}{c}{Amplitude Scaling} \\
\cline{2-7}
& $\rho \uparrow$ & IoU $\uparrow$
& $\rho \uparrow$ & IoU $\uparrow$
& $\rho \uparrow$ & IoU $\uparrow$ \\
\hline
MCC-E & 0.9839 & 0.8618 & 0.9436 & 0.6635 & 0.9827 & 0.8402 \\
MCC-H & 0.9935 & 0.9056 & 0.9691 & 0.7797 & 0.9849 & 0.8093 \\
MTC-E & 0.9824 & 0.8461 & 0.9468 & 0.6493 & 0.9805 & 0.8226 \\
MTC-H & 0.9943 & 0.9085 & 0.9831 & 0.7976 & 0.9929 & 0.8948 \\
\hline
\textbf{Average}
& \textbf{0.9885} & \textbf{0.8805}
& \textbf{0.9607} & \textbf{0.7225}
& \textbf{0.9853} & \textbf{0.8417} \\
\hline
\end{tabular}
\end{table*}

As shown in Table~\ref{tab:robustness}, CoLaX exhibits high explanation consistency across all four datasets. Gaussian noise and amplitude scaling yield average Spearman correlations of 0.9885 and 0.9853, respectively, indicating stable saliency rankings. Temporal shifts produce relatively lower overlap (IoU = 0.7225), potentially reflecting changes in the positions of salient patterns. Nevertheless, the average rank correlation remains high (0.9607), suggesting that CoLaX maintains substantial explanation consistency under the evaluated prediction-preserving perturbations.

\subsection{Effect of Context-dependent Prototype Weighting}
\label{appendix-static-weights}

To examine whether the benefits of CoLaX arise from context-dependent weighting rather than merely introducing learnable prototype weights, we introduce a \textit{Static Learnable Weights} variant. This variant removes the global encoder, context bank, global alignment loss, and context-to-prototype interaction, replacing them with an input-independent learnable weight vector $\boldsymbol{w}_{\mathrm{static}} \in \mathbb{R}^{K_{\mathrm{local}}}$. The local prototype learning, segment extraction protocol,
perturbation operator, and black-box Shapley estimator are retained. However, the context-dependent modulation is replaced by static prototype weighting.

\begin{table}[t]
\centering
\caption{Comparison between CoLaX and static learnable prototype weights. Best results are highlighted in bold.}
\label{tab:static-weights}
\scriptsize
\begin{tabular}{llccc}
\hline
Dataset & Method & AUPRC $\uparrow$ & AUP $\uparrow$ & AUR $\uparrow$ \\
\hline
\multirow{2}{*}{MCC-E}
& CoLaX & \textbf{0.933$\pm$0.003} & \textbf{0.764$\pm$0.004} & \textbf{0.853$\pm$0.003} \\
& Static Weights & 0.284$\pm$0.003 & 0.293$\pm$0.004 & 0.268$\pm$0.004 \\
\hline
\multirow{2}{*}{MCC-H}
& CoLaX & \textbf{0.977$\pm$0.000} & \textbf{0.904$\pm$0.002} & \textbf{0.769$\pm$0.005} \\
& Static Weights & 0.785$\pm$0.002 & 0.798$\pm$0.003 & 0.671$\pm$0.005 \\
\hline
\multirow{2}{*}{MTC-E}
& CoLaX & \textbf{0.723$\pm$0.007} & \textbf{0.638$\pm$0.007} & \textbf{0.784$\pm$0.006} \\
& Static Weights & 0.660$\pm$0.006 & 0.595$\pm$0.005 & 0.762$\pm$0.005 \\
\hline
\multirow{2}{*}{MTC-H}
& CoLaX & \textbf{0.920$\pm$0.001} & \textbf{0.686$\pm$0.001} & 0.977$\pm$0.000 \\
& Static Weights & 0.819$\pm$0.004 & 0.621$\pm$0.003 & \textbf{0.986$\pm$0.001} \\
\hline
\end{tabular}
\end{table}

As shown in Table~\ref{tab:static-weights}, CoLaX outperforms the static-weight variant on 11 of 12 metrics, improving average AUPRC from 0.637 to 0.888, AUP from 0.577 to 0.748, and AUR from 0.672 to 0.846. Unlike static weights, which assign identical prototype relevance across samples, CoLaX adapts prototype weights to the global temporal context of each input. These results highlight the importance of global context conditioning for constructing more informative explanatory segments, beyond simply introducing learnable prototype weights.

\subsection{More experiments on Real-world dataset}\label{ucr_more}
\begin{figure*}
    \centering
    \includegraphics[width=0.8\linewidth]{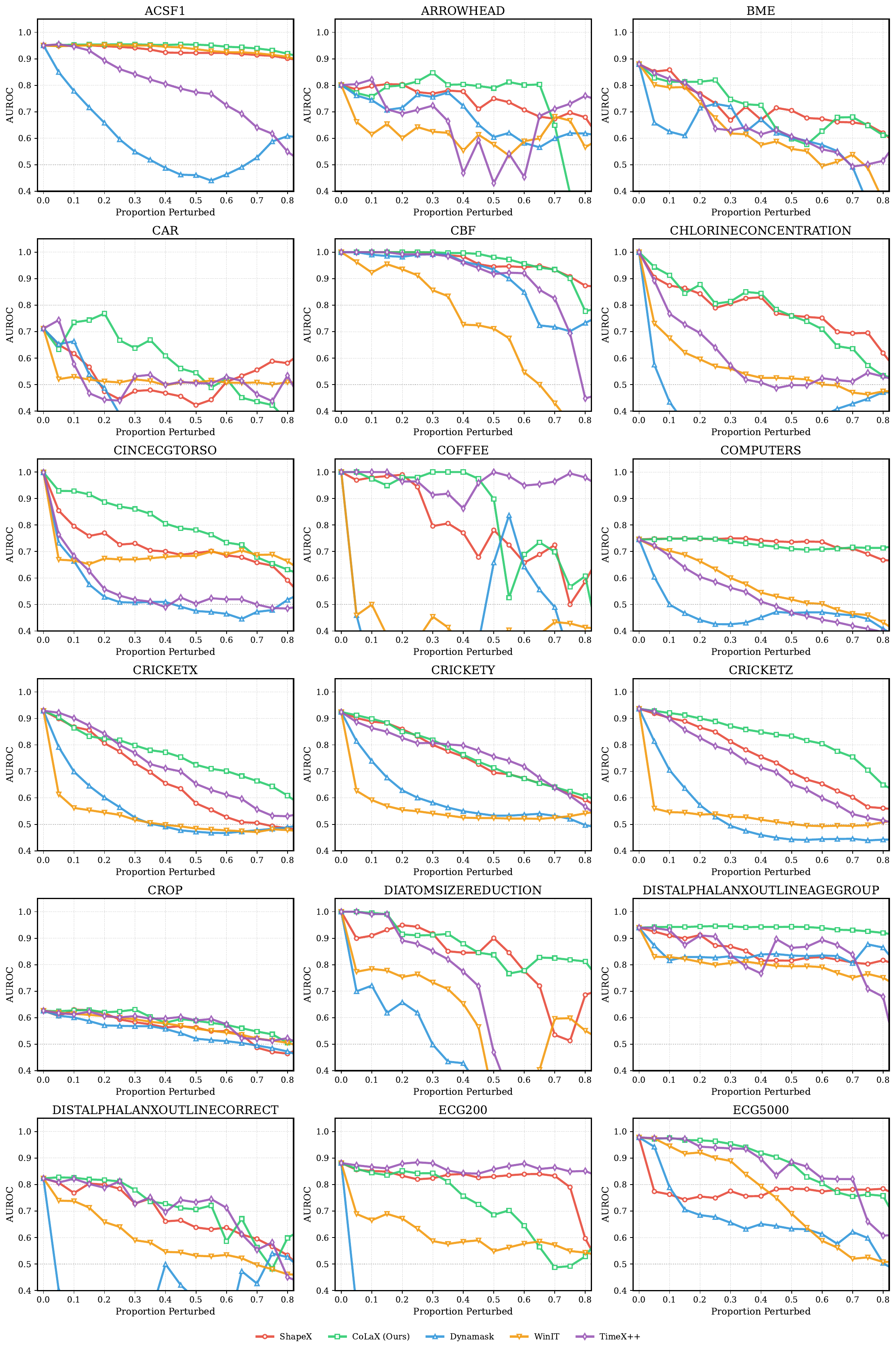}
    \caption{The occlusion experimental results on UCR archive (1/2).}
    \label{ucr_all1}
\end{figure*}

\begin{figure*}
    \centering
    \includegraphics[width=0.8\linewidth]{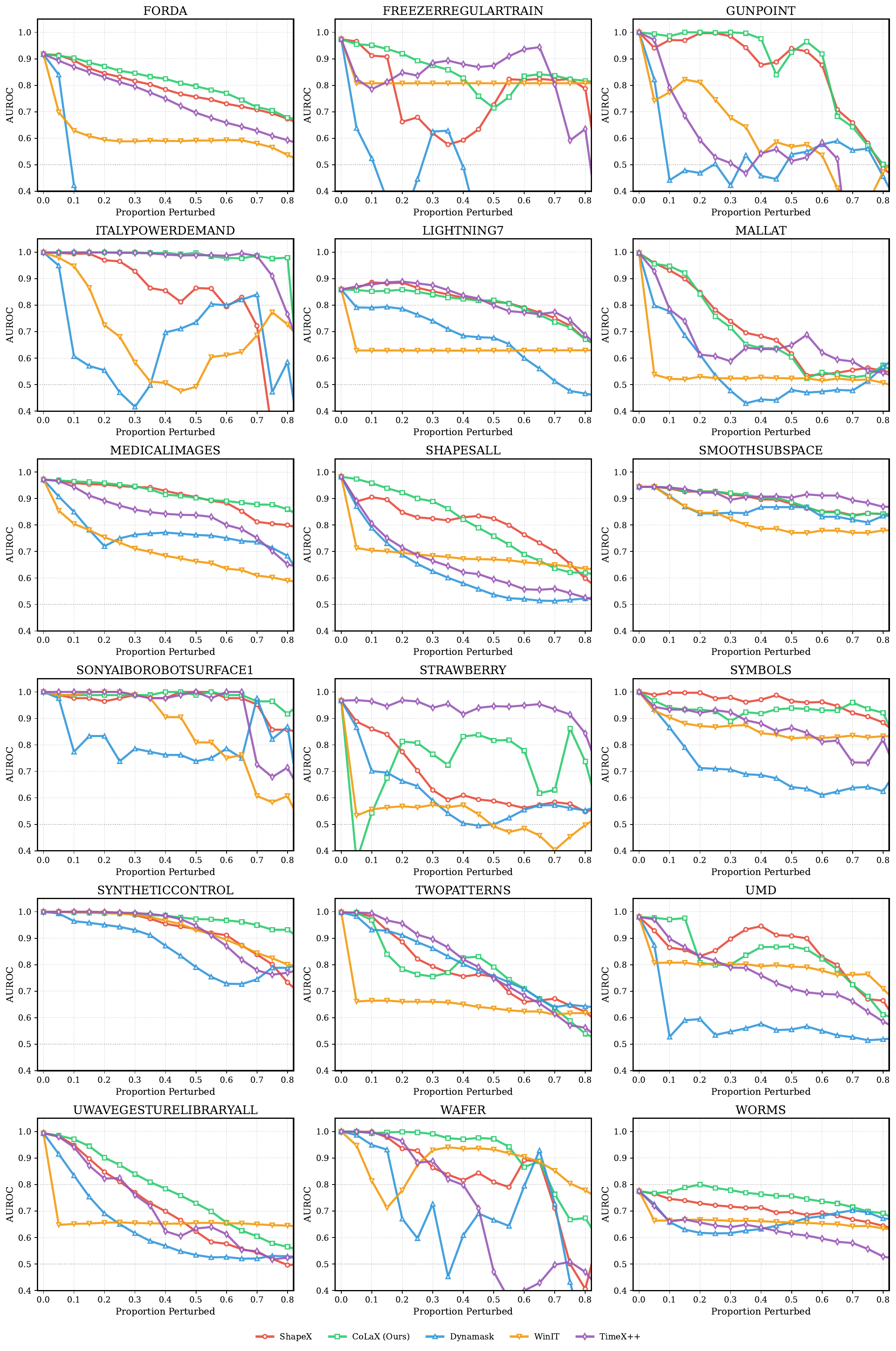}
    \caption{The occlusion experimental results on UCR archive (2/2).}
    \label{ucr_all2}
\end{figure*}

The \textcolor{black}{UCR Archive} is widely adopted for comparison given their inherent diversity, covering various domains, sequence lengths, and class cardinalities. The results for occlusion-based evaluation are illustrated in Figure \ref{ucr_all1} and \ref{ucr_all2}.

It can be inferred that no single explanation model dominates across all datasets. Our CoLaX appears to degrade more gracefully on challenging cases, as reflected in the smoother drop of AUROC scores across datasets, illustrating its robustness.
\subsection{Interpretable comparisons} \label{appendix-ecg}
\begin{figure*}
    \centering
    \includegraphics[width=0.9\linewidth]{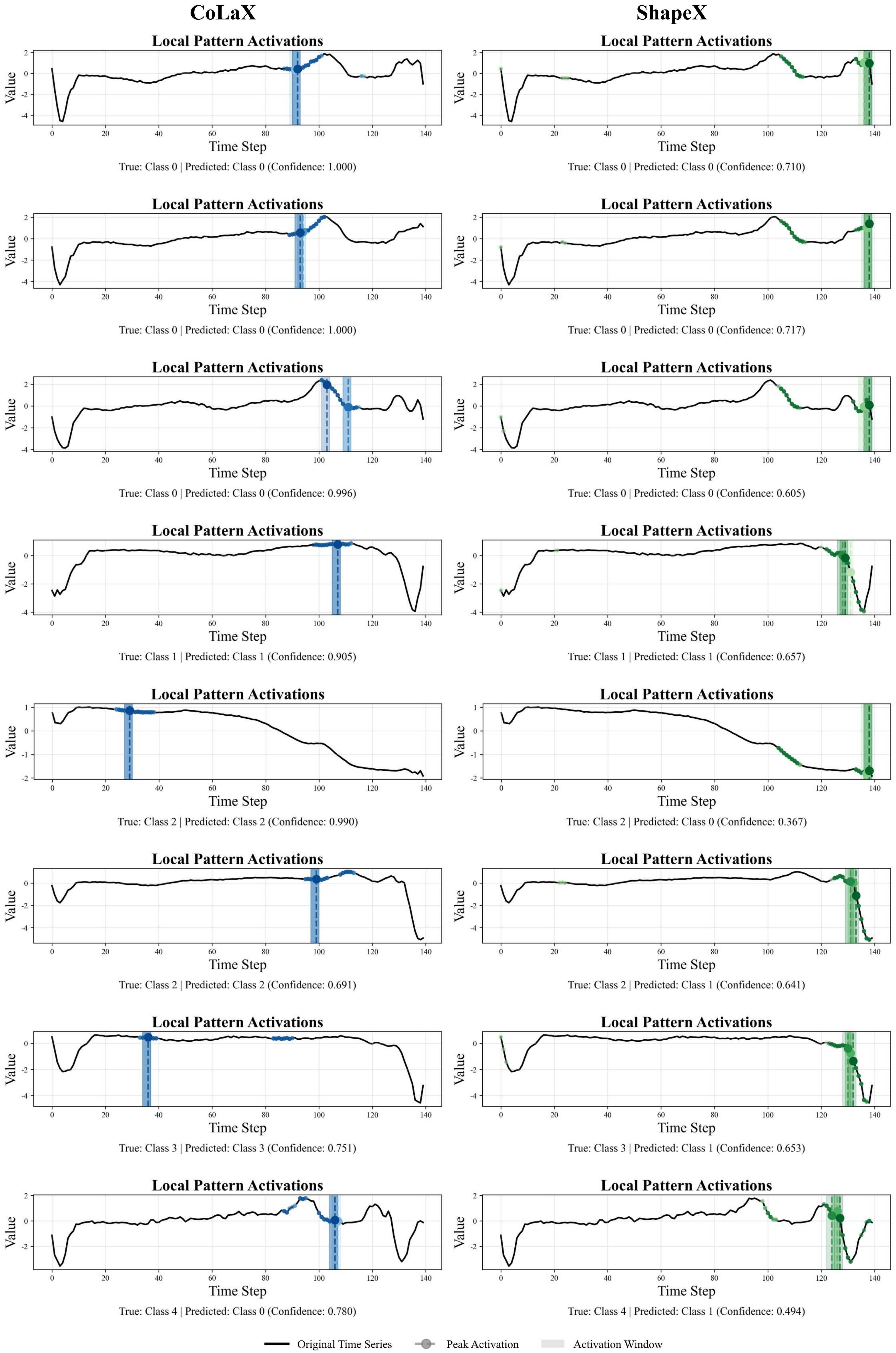}
    \caption{Comparison of Class-Discriminative Regions learned by CoLaX and ShapeX on the ECG5000 Dataset. Predictions and confidence scores refer to the outputs of the internal surrogate heads used to validate the representation fidelity of the learned prototype dictionary.}
    \label{ecg5000}
\end{figure*}

To evaluate whether the models capture genuine electrophysiological features during their initial training phase, we visualize the internal surrogate activations of CoLaX versus ShapeX on the ECG5000 dataset (Figure \ref{ecg5000}). This comparison reveals a fundamental divergence in the internal representation logic learned by the two frameworks.

The ECG5000 benchmark, derived from the BIDMC Congestive Heart Failure Database (CHFDB) \cite{physiobank2000physionet}, comprises 5,000 heartbeat segments. Originally sampled at 250 Hz, each segment was length-normalized to 140 time steps via interpolation; thus, the temporal axis reflects the relative cardiac phase rather than absolute duration. The dataset includes five heartbeat classes: Normal (N), R-on-T PVC, PVC, SP/EB, and Unclassified (UB). We report the surrogate performance of both models across these classes to validate the quality of their learned prototype dictionaries.

It can be clearly inferred from Figure \ref{ecg5000} that CoLaX demonstrates dynamic adaptability, with activation regions (blue) shifting distinctively according to the morphological characteristics of each class. Conversely, ShapeX exhibits a static bias, with its activations (green) consistently fixated on the tail of the time series (Time Step $>$ 130), regardless of the input pathology. Specifically, for Normal and R-on-T classes, CoLaX concentrates on the T-wave repolarization phase (Time Step $\sim$90--110), whereas for PVCs, it shifts attention to the onset QRS complex. However, ShapeX exhibits a static bias, focusing more on the tail of the signal. As previously discussed, the dataset underwent interpolation and length normalization, enforcing a fixed resolution of 140 time steps for all heartbeats. A critical artifact of this process lies at the boundaries of the time series, particularly the tail. These regions often accumulate significant non-biological information, such as edge-smoothing artifacts introduced by interpolation algorithms or artificial zero-padding effects resulting from segmentation. These artifacts act as spurious features. The fact that CoLaX avoids these regions suggests that its representation learning is more robust and physiologically grounded.

In particular, with proper accuracy and confidence, CoLaX focuses more on the characteristic of the given signal. This difference is particularly evident in the Class 2 sample. CoLaX's internal surrogate head captures physiologically consistent features, whereas ShapeX’s representation may reflect boundary-related artifacts. Consequently, ShapeX’s surrogate prediction fails to align with the actual class, categorizing the severe arrhythmia as 'Normal' with low surrogate confidence ($0.367$). We emphasize that this does not directly measure post-hoc faithfulness to a black-box; rather, it demonstrates that CoLaX learns a higher-fidelity dictionary. 

It is worth noting that both surrogates struggle with Class 4 identification. Unlike ventricular arrhythmias (e.g., Class 2), which exhibit distinct broad QRS complexes, Class 4 waveforms often present with subtle morphological deviations that significantly overlap with normal beats. Consequently, improving the sensitivity to Class 4 through fine-grained feature extraction and multi-scale attention mechanisms remains a priority for our future research.

\subsection{Additional Comparisons with Subsequence-Based Explainers}
\label{app:additional_baselines}

To further assess the fairness of baseline comparisons, we include two additional methods: Implet~\cite{meng2025implet}, a post-hoc subsequence-based explainer, and TIMING~\cite{jang2025timing}, a temporality-aware Integrated Gradients method. These methods complement the existing baselines by providing more structurally relevant comparisons beyond conventional point-wise attribution.

Table~\ref{tab:additional_baselines} reports the results on four synthetic benchmarks. CoLaX consistently outperforms Implet across all datasets and evaluation metrics. Compared with TIMING, CoLaX achieves higher AUPRC and AUR on all four datasets, while TIMING obtains a higher AUP on MTC-H. These results suggest that the improvements of CoLaX cannot be attributed solely to comparisons with conventional point-wise explanation methods.

\begin{table}[t]
\centering
\caption{Additional comparisons with Implet and TIMING on four synthetic datasets. The best results are highlighted in bold.}
\label{tab:additional_baselines}
\scriptsize
\begin{tabular}{llccc}
\toprule
Dataset & Method & AUPRC & AUP & AUR \\
\midrule
\multirow{3}{*}{MCC-E}
& CoLaX & $\mathbf{0.933\pm0.003}$ & $\mathbf{0.764\pm0.004}$ & $\mathbf{0.853\pm0.003}$ \\
& Implet & $0.382\pm0.005$ & $0.522\pm0.003$ & $0.501\pm0.003$ \\
& TIMING & $0.364\pm0.005$ & $0.472\pm0.008$ & $0.469\pm0.005$ \\
\midrule
\multirow{3}{*}{MCC-H}
& CoLaX & $\mathbf{0.977\pm0.000}$ & $\mathbf{0.904\pm0.002}$ & $\mathbf{0.769\pm0.005}$ \\
& Implet & $0.525\pm0.006$ & $0.488\pm0.004$ & $0.556\pm0.004$ \\
& TIMING & $0.446\pm0.007$ & $0.582\pm0.009$ & $0.433\pm0.008$ \\
\midrule
\multirow{3}{*}{MTC-E}
& CoLaX & $\mathbf{0.723\pm0.007}$ & $\mathbf{0.638\pm0.007}$ & $\mathbf{0.784\pm0.006}$ \\
& Implet & $0.174\pm0.002$ & $0.180\pm0.001$ & $0.508\pm0.001$ \\
& TIMING & $0.188\pm0.002$ & $0.449\pm0.007$ & $0.412\pm0.005$ \\
\midrule
\multirow{3}{*}{MTC-H}
& CoLaX & $\mathbf{0.920\pm0.001}$ & $0.686\pm0.001$ & $\mathbf{0.977\pm0.000}$ \\
& Implet & $0.431\pm0.006$ & $0.523\pm0.001$ & $0.458\pm0.005$ \\
& TIMING & $0.402\pm0.005$ & $\mathbf{0.799\pm0.007}$ & $0.206\pm0.005$ \\
\bottomrule
\end{tabular}
\end{table}

Nevertheless, these synthetic datasets are deliberately constructed around discriminative temporal motifs and may favor methods that produce coherent subsequence-level explanations. We therefore place particular emphasis on comparisons with structurally related methods, especially ShapeX and Implet. The results support the effectiveness of CoLaX in motif-rich settings, particularly when the relevance of recurring local patterns depends on the broader temporal context.

\subsection{Visualizations for local pattern and global context}
\begin{figure*}[t]
    \centering
    \begin{subfigure}{0.48\linewidth}
        \centering
        \includegraphics[width=\linewidth]{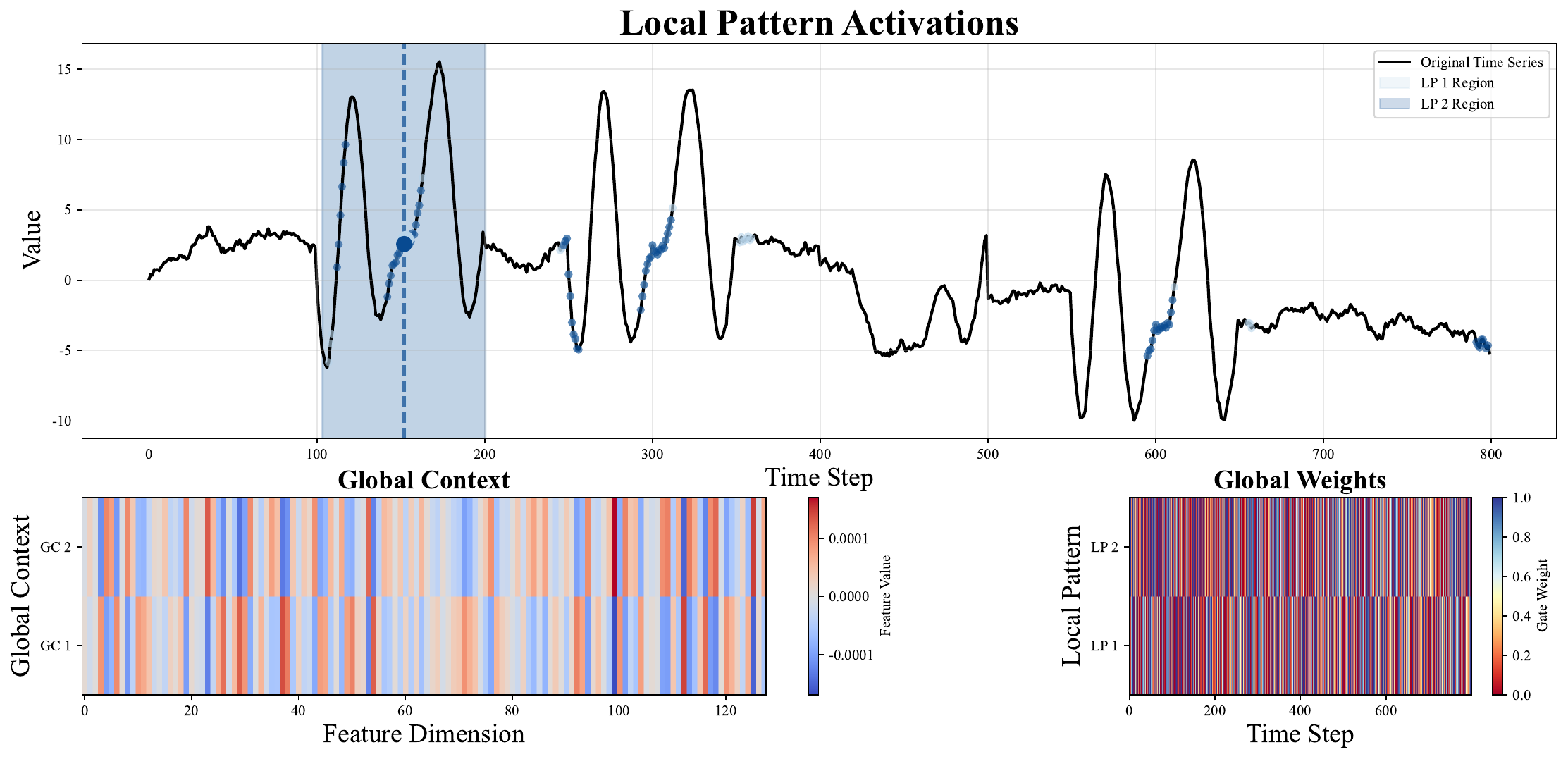}
        \caption{MCCE}
        \label{fig:mcce_vis}
    \end{subfigure}
    \hfill
    \begin{subfigure}{0.48\linewidth}
        \centering
        \includegraphics[width=\linewidth]{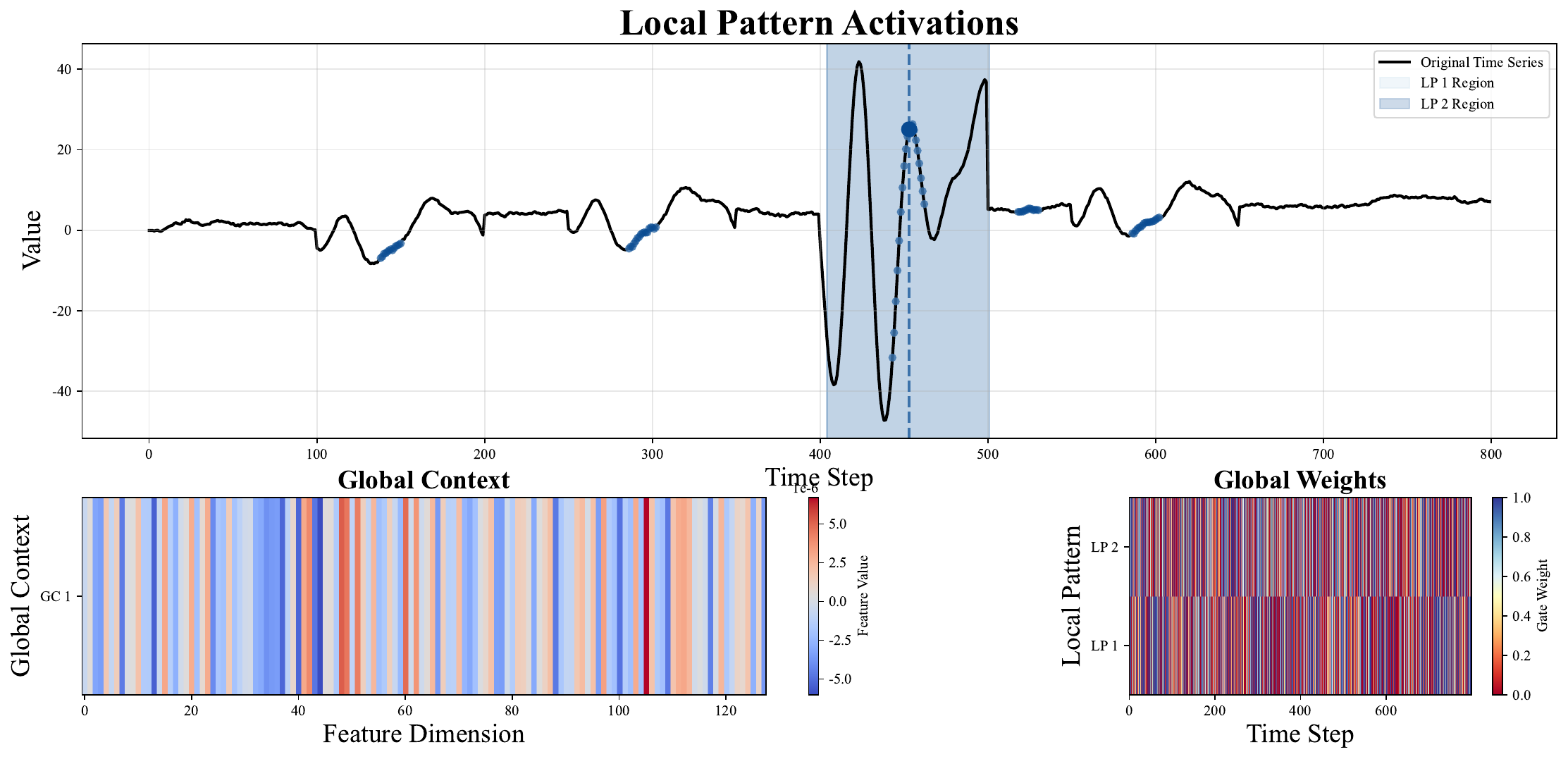}
        \caption{MCCH}
        \label{fig:mcch_vis}
    \end{subfigure}
    \caption{Visualization of model explanations on synthetic datasets.
    Saliency maps highlight the localized discriminative segments identified by the model under different motif–context configurations.}
    \label{synthetic_vis}
\end{figure*}

Figure \ref{synthetic_vis} provides qualitative insights into how local pattern activations, global context representations, and their context-dependent gating jointly contribute to the model’s decision-making process.

The top panel displays the original time series together with multiple detected local pattern activation regions. These regions correspond to temporally localized fluctuations with distinct morphological characteristics. Notably, similar local patterns appear at different time intervals, including both pronounced transient events and more subtle variations. This observation highlights that the mere occurrence of a local pattern does not uniquely determine its predictive relevance.

The bottom-left panel visualizes the learned global context representations, which capture slowly varying, high-level temporal structures across feature dimensions. These global context embeddings evolve smoothly over time and are not temporally aligned with individual local activations, indicating that they encode background system conditions rather than instantaneous dynamics.

Crucially, the bottom-right panel illustrates the resulting global-local gating weights, which modulate the contribution of each local pattern over time. The sample-specific gating weights adjust the overall relevance of each local prototype, while the temporal context signal modulates its activation strength across time. Consequently, context-conditioned activations can vary across temporal locations even when the underlying prototype remains unchanged. In particular, strong local transients receive high gating weights only when aligned with compatible global context states, whereas similar local patterns occurring under different global conditions are down-weighted.

Overall, these visualizations provide intuitive evidence that our CoLaX does not rely on local patterns in isolation. Instead, it implements a context-aware weighting mechanism in which global temporal context conditions the contribution of local patterns to the final prediction. This qualitative behavior is consistent with the non-additive and interaction-driven trends observed in the quantitative parameter analysis and ablation studies, and supports the interpretability of the proposed global-to-local modulation design.
\section{The Saliency Score for CoLaX} \label{appendix-sal}

\subsection{Definition of Saliency Score}

Given a trained classification model $f$ that maps an input time series $\boldsymbol{X}$ to a predicted label $\hat{Y}$, CoLaX aims to identify the contribution of shapelet-aligned subsequences to the model's decision. Unlike timestep-level attribution, we focus on identifying meaningful segments driven by learned shapelets. A saliency map $R$ is generated to quantify this importance, adopting the methodology proposed in ShapeX \cite{huang2025shapex}.

Let $\boldsymbol{X}=[x_1, x_2, \cdots, x_T]$ denote a time series of length $T$. The saliency map $R$ is defined as a sequence of importance values, where each element $R_t$ quantifies the contribution of $x_t$ to the model's prediction $f(\boldsymbol{X})$. Formally, Shapley values are employed to quantify the marginal contribution of each segment to the prediction $f(\boldsymbol{X})$. The final saliency score $R_t$ for each time step corresponds to the importance of the shapelet-driven segment to which it belongs. The normalized saliency score $R_t$ represents the magnitude of the aggregated segment-level attribution at time step $t$. Higher values indicate greater attribution magnitude, without distinguishing positive from negative contributions. For visualization and comparison purposes, saliency scores are normalized to the range $[0, 1]$.

\subsection{Perturbation-based Explanation Methods}
Adopting the perturbation-based paradigm, the saliency score is interpreted as the expected variation in the model's output under input perturbation. Let $P(\boldsymbol{X}, \boldsymbol{M})$ denote a perturbation function that modifies the input $\boldsymbol{X}$ based on a binary mask $\boldsymbol{M} \in \{0, 1\}^T$. The perturbed input $\boldsymbol{X}'$ is defined as:
\begin{equation}
    \boldsymbol{X}' = \boldsymbol{M} \odot \boldsymbol{X} + (1 - \boldsymbol{M}) \odot \boldsymbol{X}_{\text{base}},
\end{equation}
where $\odot$ denotes the Hadamard product with $\boldsymbol{M}$ broadcasted across the $D$ feature dimensions. Here, $\boldsymbol{X}_{\text{base}}$ represents a context-consistent baseline signal generated via linear interpolation to maintain temporal continuity and avoid artificial spikes at segment boundaries.

The temporal saliency map is obtained by aggregating the absolute segment-level Shapley values:
\begin{equation}
\boldsymbol{R} =
\operatorname{Aggregate}
\left(\{\phi_n,\mathcal{I}_n\}_{n=1}^{N}\right),
\end{equation}
where $\operatorname{Aggregate}(\cdot)$ denotes the maximum absolute aggregation and normalization defined in Eq.~\ref{eq-final-aggregation}.

\subsection{Visualization of Saliency Maps}

\begin{figure*}
    \centering
    \includegraphics[width=1\linewidth]{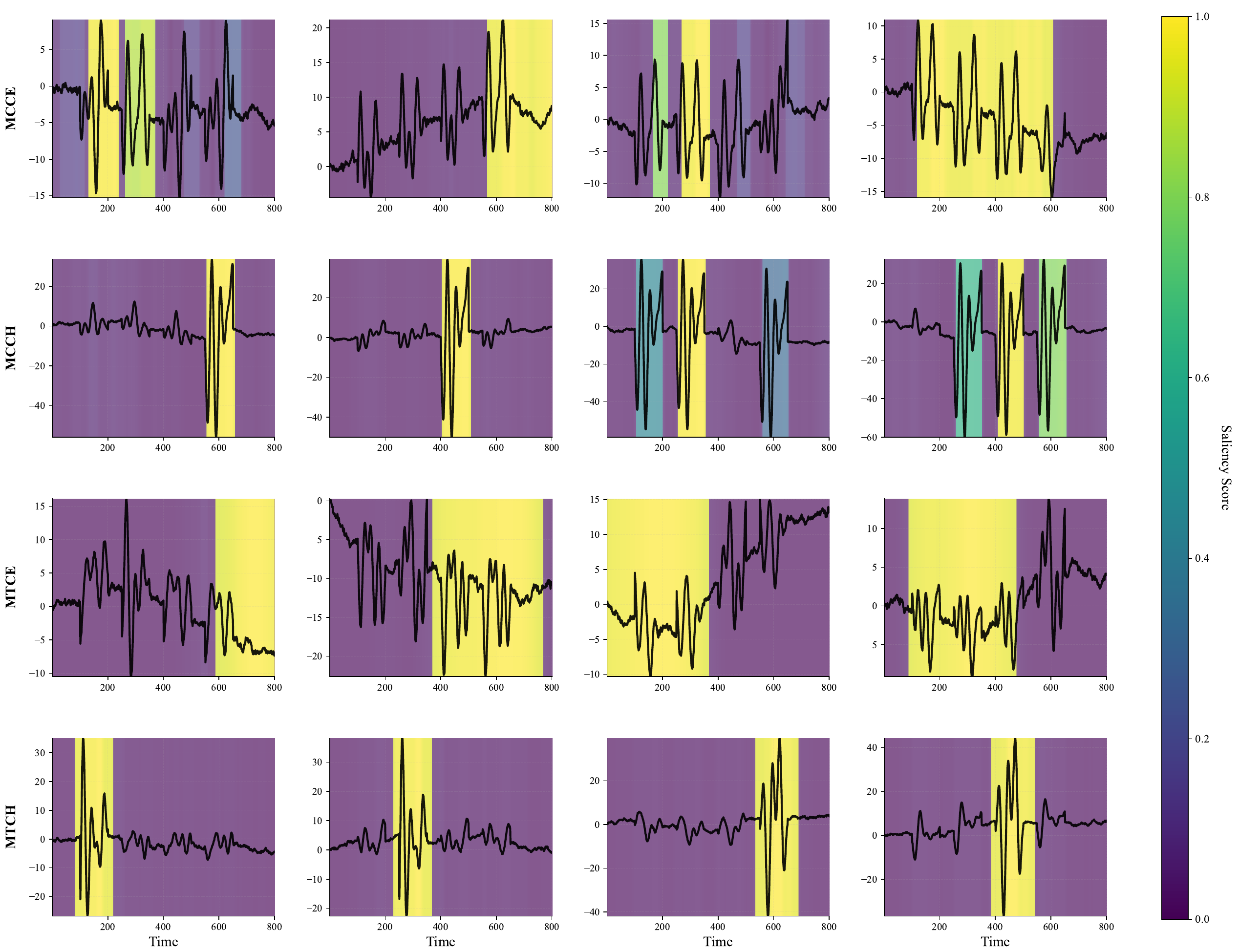}
    \caption{The visualization of saliency score in MCC-E, MCC-H, MTC-E, and MTC-H dataset.}
    \label{vis-sal}
\end{figure*}

Figure \ref{vis-sal} presents the saliency maps generated by CoLaX on the four synthetic datasets.
{\color{black}

\section{Conceptual Motivation for Context-Aware Attribution}
This section provides an intuitive motivation for context-aware attribution rather than a formal theoretical guarantee. From a conditional-attribution perspective, the relevance of a local temporal pattern may depend on the broader sequence-level context in which it occurs.

\paragraph{Conceptual Analogy to Conditional Attribution.}
Consider a time series characterized by local temporal patterns $X_{\mathrm{local}}$ and global context $X_{\mathrm{global}}$. The marginal relationship $P(Y\mid X_{\mathrm{local}})$ does not explicitly distinguish the different contexts in which a local pattern may appear. In contrast, $P(Y\mid X_{\mathrm{local}},X_{\mathrm{global}})$ illustrates how the predictive relevance of the same local pattern may vary across global temporal conditions. This distinction motivates incorporating global context when constructing local explanations, without implying that these conditional distributions are
explicitly estimated or that the resulting attributions have a causal interpretation.

\textbf{Context-aware Representation and Attribution.} Motivated by this perspective, CoLaX learns a global representation $\boldsymbol{E}_{\mathrm{global}}$ to generate sample-specific prototype weights $\boldsymbol{w}$. During training, these weights modulate the learned prototype representations through $\widetilde{\boldsymbol{Q}}=\operatorname{diag}(\boldsymbol{w})\widetilde{\boldsymbol{P}}$. During post-hoc inference, the same weights, together with the temporal context signal $c_t$, modulate local activations as $A^{\mathrm{ctx}}_{t,k}=w_k A_{t,k}(1+\alpha c_t)$. The resulting activations determine the candidate segments used for black-box Shapley attribution. Thus, global context influences which local patterns are selected as explanatory units, while
their contributions are evaluated through the frozen classifier's prediction changes. This design provides an explicit mechanism for context-dependent attribution, rather than a theoretical guarantee of explanation faithfulness.}

\section{Algorithm for CoLaX}\label{appendix-alg}
The training and inference procedures of CoLaX are summarized in Algorithm \ref{algorithm:colax}:

\begin{algorithm}[!htbp]
\caption{Training and Inference of \textbf{CoLaX}}
\label{algorithm:colax}
\small  % 或 \small
\begin{algorithmic}[1]
\REQUIRE Multivariate time series $\boldsymbol{X} \in \mathbb{R}^{T \times D}$, labels $\boldsymbol{y} \in \{0,1\}^C$, frozen black-box classifier $f(\cdot)$;
local prototype length $L_p$, number of prototypes $K_{\text{local}}$, number of global contexts $K_{\text{global}}$, temperatures $\tau_1,\tau_2$, modulation strength $\alpha$, activation threshold $\Omega$, loss weights $\lambda_1,\lambda_2$, learning rate $\eta$.
\REQUIRE CoLaX parameters $\theta$, local prototypes $\boldsymbol{P}$, and global context bank $\mathcal{G}$.

\STATE \textbf{Training Phase (CoLaX modules trainable)}
\FOR{epoch $= 1$ to max epoch}
    \FOR{each batch of input data $(\boldsymbol{X}, \boldsymbol{y})$}
        \STATE \textbf{Local Pattern Branch:}
        \STATE Refine local prototypes: $\widetilde{\boldsymbol{P}} = \text{CTEncoder}(\boldsymbol{P})$
        \STATE Compute activation map: $\boldsymbol{A} = \text{Softmax}(\text{Conv}(\boldsymbol{X}, \widetilde{\boldsymbol{P}}))$ 
        
        \STATE \textbf{Global Context Branch:}
        \STATE Extract global embeddings: $\boldsymbol{Z} = \text{GEncoder}(\boldsymbol{X})$
        \STATE Compute context attention: $a_{t,k} = \text{Softmax}(\boldsymbol{z}_t^\top \boldsymbol{g}_k / \tau_1)$
        \STATE Aggregate global representation: $\boldsymbol{E}_{\text{global}} = \text{Pool}(\boldsymbol{Z})$
        
        \STATE \textbf{Context-Aware Modulation:}
        \STATE Query $\boldsymbol{q} = \boldsymbol{W}_q \boldsymbol{E}_{\text{global}}$, Key $\boldsymbol{K} = \widetilde{\boldsymbol{P}} \boldsymbol{W}_k$
        \STATE Compute gating: $\boldsymbol{w} = \operatorname{Sigmoid} (\boldsymbol{K}\boldsymbol{q}/\tau_2)$
        \STATE Modulate local prototypes: $\widetilde{\boldsymbol{Q}} = \text{diag}(\boldsymbol{w})\widetilde{\boldsymbol{P}}$
        
        \STATE \textbf{Prediction:} $\hat{\boldsymbol{y}} = \sigma(\text{MLP}(\widetilde{\boldsymbol{Q}}))$ 
        
        \STATE \textbf{Loss Computation:}
        \STATE $\mathcal{L} = \mathcal{L}_{\text{cls}} + \lambda_1 \mathcal{L}_{\text{l-match}} + \lambda_2 \mathcal{L}_{\text{g-align}}$
        
        \STATE \textbf{Optimization:} Update $\theta$ via gradient descent
        \STATE \textbf{Memory Update:} 
        \STATE Compute normalized
    $\bar{\boldsymbol{z}}_k =
    \frac{\sum_{t=1}^{T'} a_{t,k}\boldsymbol{z}_t}
    {\sum_{j=1}^{T'} a_{j,k} + \epsilon}$
        \STATE Update $\mathcal{G}$ via gated memory rule.
    \ENDFOR
\ENDFOR

\STATE \textbf{Inference Phase (CoLaX modules frozen)}
\FOR{each test sample $\boldsymbol{X}_{\text{test}}$}
    \STATE Compute local activation map $\boldsymbol{A}$, temporal embeddings $\boldsymbol{Z}$, pooled global context $\boldsymbol{E}_{\text{global}}$, sample-specific prototype weights $\boldsymbol{w}$, and context attention $a_{t,k}$. 
    \STATE Compute temporal context signal $\boldsymbol{c}$ from the context-attention matrix:\\
    $\displaystyle
    c_t\gets\mathrm{Norm}\left[
    \mathrm{Interp}\left(
    \{\max_k a_{s,k}\}_{s=1}^{T'}
    \right)\right]_t$
    \STATE Construct context-conditioned activation $A^{\text{ctx}}_{t,k} = w_k A_{t,k}(1+\alpha c_t).$
    
    \STATE Extract contiguous intervals satisfying
    $A^{\mathrm{ctx}}_{t,k}>\Omega$.
    \STATE Form the candidate segment set
    $\mathcal{P}(\boldsymbol{X}_{\mathrm{test}})
    =\{\mathcal{I}_1,\ldots,\mathcal{I}_N\}$.
    \STATE Construct coalition-based perturbations
    using the detected segments.
    \STATE \textbf{Black-box Shapley Attribution:} $c^\star\gets
    \arg\max_c f_c(\boldsymbol{X}_{\text{test}})$
    \STATE Construct coalition perturbations
    $\boldsymbol{X}_S$ using interpolation masking.
    \STATE Evaluate $v(S)\gets f_{c^\star}(\boldsymbol{X}_S)$
    and compute segment Shapley values $\{\phi_n\}$.
    \STATE Aggregate $|\phi_n|$ over overlapping segments
    and normalize to obtain $\boldsymbol{R}$.
\ENDFOR

\RETURN Trained CoLaX parameters $\theta$ and saliency maps $\{\boldsymbol{R}\}.$

\end{algorithmic}
\end{algorithm}

\section{Limitations}
\label{limitations}

While CoLaX addresses local--global ambiguity in time-series explanations, several limitations remain.

First, as a post-hoc surrogate-based framework, CoLaX inherits the limitations of perturbation-based explainability. Its explanations characterize the responses of a frozen black-box classifier to structured input perturbations rather than revealing causal mechanisms of the underlying data-generating process. Moreover, the resulting attributions remain sensitive to the perturbation strategy, as segment-level masking may generate out-of-distribution samples. Since CoLaX is trained independently of the black-box classifier, its learned prototypes may also fail to capture classifier-specific or spurious features, limiting explanation faithfulness.

Second, the empirical advantages of CoLaX are most pronounced on synthetic, motif-rich datasets designed to exhibit local--global ambiguity. Improvements on real-world datasets are generally less consistent. Although CoLaX achieves the highest AUPRC on ECG, it does not outperform ShapeX on AUP and AUR, and no single method consistently dominates across the heterogeneous UCR datasets. Thus, the current results support its effectiveness primarily in settings where recurring local patterns have context-dependent relevance, rather than establishing universal superiority.

Third, although CoLaX supports multivariate inputs and has been evaluated on several multivariate benchmarks, the experimental evaluation remains predominantly univariate. Its ability to capture complex cross-variable interactions in high-dimensional settings requires further investigation.

Fourth, the dual-branch architecture and representation-learning stage introduce additional computational overhead compared with simpler attribution methods. While computationally practical on the evaluated benchmarks, scalability to extremely long sequences and high-dimensional multivariate data remains an open challenge.

Finally, CoLaX remains a post-hoc explanation framework without formal guarantees of attribution faithfulness or stability. Extending it toward inherently interpretable architectures and establishing theoretical guarantees under non-stationary conditions are promising directions for future research.

\section{Broader Impacts}\label{broader}
CoLaX helps practitioners inspect time-series classifiers by highlighting which local patterns a frozen model responds to and in what sequence-level contexts, supporting debugging, spurious-cue detection, and expert auditing (e.g., healthcare or environmental monitoring).

CoLaX provides post-hoc, correlational approximations of a model’s response to structured perturbations; results depend on the intervention strategy and prototype dictionary, and they do not establish causality or clinical validity. Benefits are strongest on synthetic benchmarks, and real-world utility needs further study. The main risk is over-reliance (treating explanations as evidence a prediction is correct), so CoLaX should be used only as decision support with domain-specific validation and human oversight, not for autonomous high-stakes decisions.

All experiments use public, de-identified benchmarks; we collected no new data and conducted no user study or crowd annotation. Since CoLaX explains existing models and generates no content, we see no meaningful additional dual-use risk beyond that of the underlying classifiers.

%%%%%%%%%%%%%%%%%%%%%%%%%%%%%%%%%%%%%%%%%%%%%%%%%%%%%%%%%%%%

\newpage
\section*{NeurIPS Paper Checklist}

\begin{enumerate}

\item {\bf Claims}
    \item[] Question: Do the main claims made in the abstract and introduction accurately reflect the paper's contributions and scope?
    \item[] Answer: \answerYes{} % Replace by \answerYes{}, \answerNo{}, or \answerNA{}.
    \item[] Justification: The main claims presented in the abstract and introduction are well-aligned with the paper’s actual contributions. The abstract accurately outlines the core method CoLaX for time series classification, highlighting its ability to capture both local discriminative patterns and global contextual information. These core ideas are consistently developed throughout the paper (Section 2), and the experimental results on UCR datasets (Section 3) and visualizations (ECG5000, synthetic data) empirically support the stated claims.

\item {\bf Limitations}
    \item[] Question: Does the paper discuss the limitations of the work performed by the authors?
    \item[] Answer: \answerYes{} % Replace by \answerYes{}, \answerNo{}, or \answerNA{}.
    \item[] Justification: The paper includes a dedicated section in Appendix \ref{limitations} discussing the limitations of the proposed method.

\item {\bf Theory assumptions and proofs}
    \item[] Question: For each theoretical result, does the paper provide the full set of assumptions and a complete (and correct) proof?
    \item[] Answer: \answerNA{} % Replace by \answerYes{}, \answerNo{}, or \answerNA{}.
    \item[] Justification: This paper does not contain formal theoretical results such as theorems or lemmas. The contributions are primarily algorithmic and empirical, presenting the CoLaX method with experimental validation on benchmark datasets. Therefore, this question is not applicable.

    \item {\bf Experimental result reproducibility}
    \item[] Question: Does the paper fully disclose all the information needed to reproduce the main experimental results of the paper to the extent that it affects the main claims and/or conclusions of the paper (regardless of whether the code and data are provided or not)?
    \item[] Answer: \answerYes{} % Replace by \answerYes{}, \answerNo{}, or \answerNA{}.
    \item[] Justification: We have disclosed all necessary information to reproduce our main results. This includes the complete mathematical derivation of the algorithm, detailed architecture diagrams, pseudocode in Appendix \ref{appendix-alg}, and comprehensive hyperparameter settings. Furthermore, the source code and reproduction scripts provided at \url{https://github.com/zkkkkkkkkkkx/CoLaX} facilitate full replication.

\item {\bf Open access to data and code}
    \item[] Question: Does the paper provide open access to the data and code, with sufficient instructions to faithfully reproduce the main experimental results, as described in supplemental material?
    \item[] Answer: \answerYes{} % Replace by \answerYes{}, \answerNo{}, or \answerNA{}.
    \item[] Justification: The complete code for our proposed method, along with scripts to reproduce the main experimental results, is provided in the supplementary material and at \url{https://github.com/zkkkkkkkkkkx/CoLaX}.

\item {\bf Experimental setting/details}
    \item[] Question: Does the paper specify all the training and test details (e.g., data splits, hyperparameters, how they were chosen, type of optimizer) necessary to understand the results?
    \item[] Answer: \answerYes{} % Replace by \answerYes{}, \answerNo{}, or \answerNA{}.
    \item[] Justification: We specify all essential experimental details, including data splits, hyperparameter configurations in Appendix \ref{setting}.

\item {\bf Experiment statistical significance}
    \item[] Question: Does the paper report error bars suitably and correctly defined or other appropriate information about the statistical significance of the experiments?
    \item[] Answer: \answerYes{} % Replace by \answerYes{}, \answerNo{}, or \answerNA{}.
    \item[] Justification: We report the mean and standard deviation for all main experimental results across five independent runs with different random seeds.

\item {\bf Experiments compute resources}
    \item[] Question: For each experiment, does the paper provide sufficient information on the computer resources (type of compute workers, memory, time of execution) needed to reproduce the experiments?
    \item[] Answer: \answerYes{} % Replace by \answerYes{}, \answerNo{}, or \answerNA{}.
    \item[] Justification: We provide detailed information regarding the hardware specifications in Appendixes \ref{setting} and \ref{cost}.
    
\item {\bf Code of ethics}
    \item[] Question: Does the research conducted in the paper conform, in every respect, with the NeurIPS Code of Ethics \url{https://neurips.cc/public/EthicsGuidelines}?
    \item[] Answer: \answerYes{} % Replace by \answerYes{}, \answerNo{}, or \answerNA{}.
    \item[] Justification: We have carefully reviewed the NeurIPS Code of Ethics and confirm that our research, including data collection, algorithmic development, and experimental evaluation, fully adheres to all its principles and guidelines.

\item {\bf Broader impacts}
    \item[] Question: Does the paper discuss both potential positive societal impacts and negative societal impacts of the work performed?
    \item[] Answer: \answerYes{} % Replace by \answerYes{}, \answerNo{}, or \answerNA{}.
    \item[] Justification: Appendix~\ref{broader} discusses potential benefits (model auditing and debugging) and risks, chiefly over-reliance on correlational, perturbation-based explanations, and recommends use as decision support under human oversight with domain validation.
    
\item {\bf Safeguards}
    \item[] Question: Does the paper describe safeguards that have been put in place for responsible release of data or models that have a high risk for misuse (e.g., pre-trained language models, image generators, or scraped datasets)?
    \item[] Answer: \answerNA{} % Replace by \answerYes{}, \answerNo{}, or \answerNA{}.
    \item[] Justification: The models and methods developed in this research are intended for time series analysis and do not pose a high risk for misuse. Therefore, specific safeguards for responsible release are not applicable.

\item {\bf Licenses for existing assets}
    \item[] Question: Are the creators or original owners of assets (e.g., code, data, models), used in the paper, properly credited and are the license and terms of use explicitly mentioned and properly respected?
    \item[] Answer: \answerYes{} % Replace by \answerYes{}, \answerNo{}, or \answerNA{}.
    \item[] Justification: All existing assets (datasets and codebases) used in this work are properly cited.

\item {\bf New assets}
    \item[] Question: Are new assets introduced in the paper well documented and is the documentation provided alongside the assets?
    \item[] Answer: \answerYes{} % Replace by \answerYes{}, \answerNo{}, or \answerNA{}.
    \item[] Justification: We provide the full implementation of our proposed method, including the core algorithm, experimental scripts, and environment configurations, in the supplementary material.

\item {\bf Crowdsourcing and research with human subjects}
    \item[] Question: For crowdsourcing experiments and research with human subjects, does the paper include the full text of instructions given to participants and screenshots, if applicable, as well as details about compensation (if any)? 
    \item[] Answer: \answerNA{} % Replace by \answerYes{}, \answerNo{}, or \answerNA{}.
    \item[] Justification: This research does not involve any crowdsourcing experiments or research with human subjects.

\item {\bf Institutional review board (IRB) approvals or equivalent for research with human subjects}
    \item[] Question: Does the paper describe potential risks incurred by study participants, whether such risks were disclosed to the subjects, and whether Institutional Review Board (IRB) approvals (or an equivalent approval/review based on the requirements of your country or institution) were obtained?
    \item[] Answer: \answerNA{} % Replace by \answerYes{}, \answerNo{}, or \answerNA{}.
    \item[] Justification: The paper does not involve any human subjects or participant-based studies. Therefore, no Institutional Review Board (IRB) approval is required.

\item {\bf Declaration of LLM usage}
    \item[] Question: Does the paper describe the usage of LLMs if it is an important, original, or non-standard component of the core methods in this research? Note that if the LLM is used only for writing, editing, or formatting purposes and does \emph{not} impact the core methodology, scientific rigor, or originality of the research, declaration is not required.
    %this research? 
    \item[] Answer: \answerNA{} % Replace by \answerYes{}, \answerNo{}, or \answerNA{}.
    \item[] Justification: LLMs were used only for writing assistance and did not play any role in the core methodology, algorithm design, experimentation, or scientific decision-making process.

\end{enumerate}

\end{document}